\documentclass[letterpaper]{article} 
\usepackage{aaai2026}  
\usepackage{times}  
\usepackage{helvet}  
\usepackage{courier}  
\usepackage[hyphens]{url}  
\usepackage{graphicx} 
\usepackage{natbib}  
\usepackage{caption} 
\usepackage{algorithm}
\usepackage{algorithmic}
\usepackage{newfloat}
\usepackage{epsfig}
\usepackage{listings}
\usepackage{amsmath}      
\usepackage{amssymb}      
\usepackage{graphicx}     
\usepackage{bm}           
\usepackage{mathtools}    
\usepackage{booktabs}
\usepackage{multirow} 
\usepackage{xcolor}
\usepackage{placeins} 
\usepackage{colortbl} 
\usepackage{arydshln}
\usepackage{tabularx}
\usepackage{pifont}
\usepackage{newfloat}
\usepackage{listings}
\DeclareCaptionStyle{ruled}{labelfont=normalfont,labelsep=colon,strut=off} 
\floatstyle{ruled}
\newfloat{listing}{tb}{lst}{}
\floatname{listing}{Listing}
\title{VQAThinker: Exploring Generalizable and Explainable Video Quality Assessment via Reinforcement Learning}
\author{
Linhan Cao$^{1\ast}$,\ 
Wei Sun$^{2\ast \heartsuit }$ ,\ 
Weixia Zhang$^{1}$, \
Xiangyang Zhu$^{3}$, \
Jun Jia$^{1}$, \\
Kaiwei Zhang$^{1}$,\
Dandan Zhu$^{2}$, \
Guangtao Zhai$^{1}$,\ 
Xiongkuo Min$^{1\dagger}$
}

\affiliations{
    \textsuperscript{\rm 1}Shanghai Jiao Tong University\\
    \textsuperscript{\rm 2}East China Normal University\\
    \textsuperscript{\rm 3}Shanghai Artificial Intelligence Laboratory \\

}

\usepackage{bibentry}

\begin{document}

\maketitle

\begin{abstract}
Video quality assessment (VQA) aims to objectively quantify perceptual quality degradation in alignment with human visual perception. Despite recent advances, existing VQA models still suffer from two critical limitations: \textit{poor generalization to out-of-distribution (OOD) videos} and \textit{limited explainability}, which restrict their applicability in real-world scenarios. To address these challenges, we propose \textbf{VQAThinker}, a reasoning-based VQA framework that leverages large multimodal models (LMMs) with reinforcement learning to jointly model video quality understanding and scoring, emulating human perceptual decision-making. Specifically, we adopt group relative policy optimization (GRPO), a rule-guided reinforcement learning algorithm that enables reasoning over video quality under score-level supervision, and introduce three VQA-specific rewards: (1) a \textbf{bell-shaped regression reward} that increases rapidly as the prediction error decreases and becomes progressively less sensitive near the ground truth; (2) a \textbf{pairwise ranking reward} that guides the model to correctly determine the relative quality between video pairs; and (3) a \textbf{temporal consistency reward} that encourages the model to prefer temporally coherent videos over their perturbed counterparts. Extensive experiments demonstrate that VQAThinker achieves state-of-the-art performance on both in-domain and OOD VQA benchmarks, showing strong generalization for video quality scoring. Furthermore, evaluations on video quality understanding tasks validate its superiority in distortion attribution and quality description compared to existing explainable VQA models and LMMs. These findings demonstrate that reinforcement learning offers an effective pathway toward building generalizable and explainable VQA models solely with score-level supervision. 
\end{abstract}

\begin{links}
    \link{Code}{https://github.com/clh124/VQAThinker}
\end{links}

\begin{figure*}[t]
\centering
\centerline{\epsfig{figure=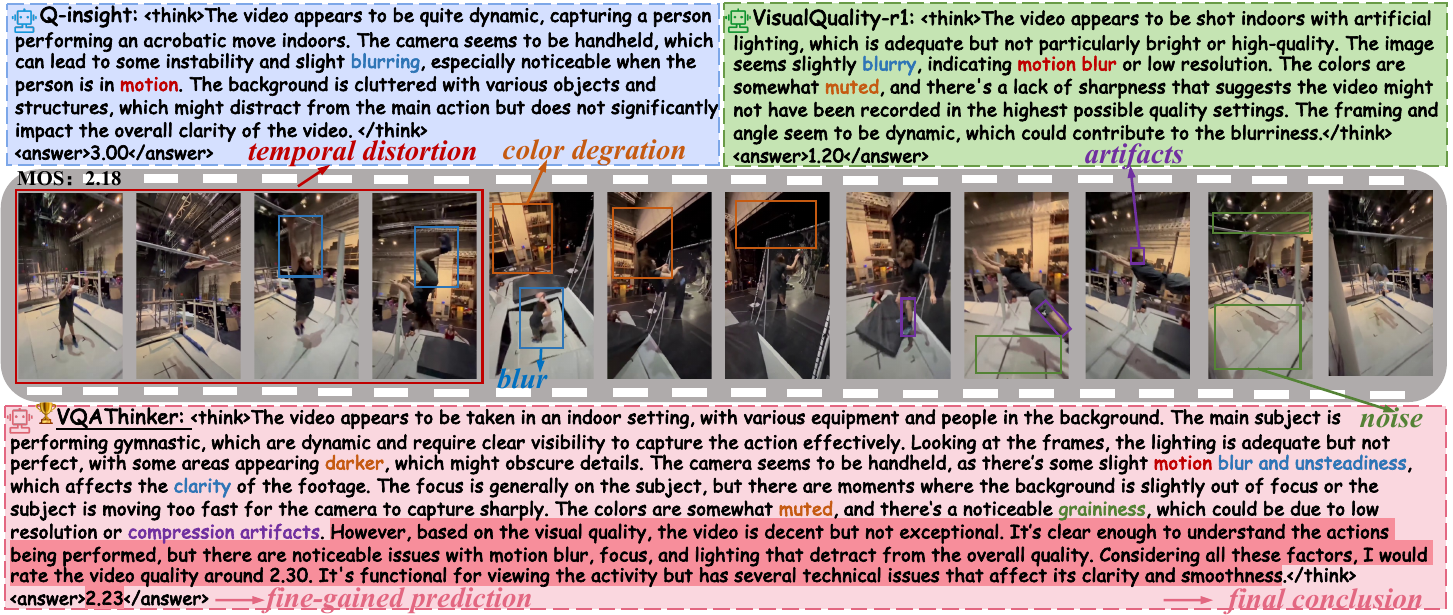,width=18cm}}
\caption{Comparison of Q-Insight, VisualQuality-R1, and our VQAThinker in video quality understanding and scoring. All three models are trained on the LSVQ dataset. Compared to Q-Insight and VisualQuality-R1, VQAThinker generates a more comprehensive reasoning trace that covers major distortion types, thereby producing more accurate video quality scores.}
\label{fig:model_structure}
\end{figure*}

\section{Introduction}

Video quality assessment (VQA)~\cite{min2024perceptual} plays a fundamental role in video processing systems, serving as the primary criterion for evaluating perceptual quality across stages such as acquisition, compression, and enhancement. Most established VQA models (\textit{e.g.,} VMAF~\cite{li2018vmaf}) rely on full-reference (FR) settings, where a pristine reference video is available for comparison. In many real-world scenarios, particularly those involving user-generated content (UGC), reference videos are unavailable, making FR methods inapplicable. Therefore, no-reference (NR) VQA, which assesses quality without relying on reference videos, offers broader applicability and is the focus of this work~\footnote{In this paper, we use VQA to refer to NR-VQA for simplicity.}.

NR-VQA has witnessed significant advancements, progressing from early handcrafted feature-based methods, such as natural scene statistics (NSS)~\cite{mittal2012no} to deep neural network (DNN) architectures, including convolutional neural networks (CNNs)~\cite{li2019quality}, Transformers~\cite{wu2022fast}, and more recently, large multimodal models (LMMs)~\cite{ge2024lmm}. However, most existing methods rely on supervised fine-tuning (SFT) using specific video content or distortion datasets~\cite{ying2021patch,li2019avc}, resulting in models that act as specialists for narrow scenarios but exhibit poor generalization to unseen video domains. To enhance generalization, recent studies have explored strategies, such as multi-dataset joint training~\cite{li2019quality}, unsupervised and self-supervised learning~\cite{cao2025breaking}, etc., aiming to improve training diversity while minimizing labeling costs. Nevertheless, these approaches often incur significantly higher training costs due to the increased data scale and model complexity.

In addition, interpretability has become a crucial capability for practical NR-VQA applications. While conventional VQA models typically output singl- or multi-dimensional quality scores, real-world scenarios often require diagnostic assessments that reveal the underlying causes of quality degradation, such as compression artifacts, temporal instability, etc. Such interpretability not only enhances the reliability of VQA predictions but also enables closed-loop video processing by guiding targeted restoration strategies based on identified distortions. Recently, the rise of LMMs has advanced the development of explainable VQA, where models are typically fine-tuned on human-annotated quality instruction datasets, such as VQA$^2$~\cite{jia2024vqa} and OmniVQA-Chat-400K~\cite{jia2025scaling}. However, these datasets are typically bootstrapped from LMM-generated descriptions and later refined by human annotators, making the final annotations susceptible to the biases and limitations of the source models. Furthermore, they often lack associated quality scores, hindering the joint learning of quality understanding and scoring, a process believed to mirror human perceptual decision-making~\cite{mazurek2003role}.

Recent advances in reinforcement learning (RL), particularly using group relative policy optimization (GRPO)~\cite{shao2024deepseekmath}, have demonstrated strong potential in enhancing the reasoning capabilities of LMMs across various tasks~\cite{li2025perception} while without the need for explicit value function. In the context of quality assessment, this learning paradigm is inherently aligned with the human perceptual process, which involves first reasoning about potential sources of quality degradation in a video and then making a quality judgment accordingly. Although GRPO has been adapted for image quality assessment (IQA)~\cite{li2025q,wu2025visualquality} and has shown improved generalizability, existing reward designs in such studies often result in coarse-grained quality predictions and lack temporal distortion modeling capabilities.

To address above challenges, we propose \textbf{VQAThinker}, a novel reasoning-driven VQA framework that leverages LMMs with reinforcement learning to jointly model video quality understanding and scoring. Specifically, we adopt GRPO to encourage the model to generate a reasoning process for video quality assessment prior to producing the final quality score, and optimize this process using score-level supervision. Under this framework, we design three VQA-specific reward functions: (1) a \textbf{bell-shaped regression reward} that enables fine-grained quality prediction by rapidly increasing rewards as the prediction error decreases and becoming progressively less sensitive near the ground truth, overcoming the limitations of constant or linearly scaled rewards in existing GRPO-based methods such as Q-Insight~\cite{li2025q} and VQ-Insight~\cite{zhang2025vq}; (2) a \textbf{pairwise ranking reward} that encourages relative quality consistency across video pairs by explicitly modeling inter-video comparison with fidelity loss~\cite{tsai2007frank}; and (3) a \textbf{temporal consistency reward} that captures temporal distortions across video frames by comparing the model’s predictions on raw and temporally perturbed videos. 

We train VQAThinker on the LSVQ dataset~\cite{ying2021patch} with score-level supervision, and experimental results show that VQAThinker not only achieves state-of-the-art performance on UGC VQA benchmarks but also outperforms existing methods on out-of-distribution (OOD) VQA benchmarks that feature diverse content and distortion types. In addition, we validate its quality understanding ability on the video distortion attribution and video quality description tasks, and the results also demonstrate that VQAThinker delivers competitive performance despite being trained without any video quality instruction data.

Our main contributions are summarized as follows:

\begin{itemize}
    \item We propose VQAThinker, a reasoning-based VQA method that combines LMMs with reinforcement learning to jointly model quality understanding and scoring.
    
    \item We design a GRPO-based training scheme with three VQA-specific rewards to enable fine-grained score regression, quality order preservation, and temporal distortion awareness under score-level supervision.

    \item Extensive experiments show that VQAThinker achieves state-of-the-art performance on both in-domain and OOD benchmarks, and delivers strong interpretability without using any instruction-tuned datasets.

\end{itemize}

\section{Related Work}
\subsection{Generalizable Video Quality Assessment}

Developing generalizable VQA models remains a long-standing objective. Early studies primarily utilize handcrafted features based on perceptual quality priors, such as NSS~\cite{mittal2012no}, motion vectors~\cite{konrad1992bayesian}, optical flow~\cite{beauchemin1995computation}, etc. While effective on synthetic distortions, these methods demonstrated limited efficacy on videos containing complex real-world distortions. Subsequently, DNN-based methods have been widely adopted for VQA, which typically comprise two main modules: a feature extraction module and a regression module. The feature extraction module often employs off-the-shelf networks, such as CNNs~\cite{li2019quality}, Transformers~\cite{wu2022fast}, and LMMs~\cite{wu2023qalign}, which are either pre-trained on large-scale datasets (\textit{e.g.}, IQA datasets~\cite{li2022blindly}) or fine-tuned directly on the target VQA datasets~\cite{sun2022deep,wu2022fast} to capture the spatial and temporal quality representation. The regression module then maps the extracted features to quality scores, using either MLPs that aggregate frame-level predictions~\cite{sun2024analysis} or sequence modeling methods, such as GRUs~\cite{li2019quality}, InceptionTime~\cite{ying2021patch}, or Transformers~\cite{wu2023discovqa}, to capture temporal dependencies across frames. DNN-based methods significantly improve the performance on in-the-wild distortions, but they still suffer from limited generalization to unseen video content and distortion types~\cite{cao2025breaking}.

To improve generalization, recent studies have explored various training strategies beyond standard supervised learning. These include mixed-dataset training with scale alignment~\cite{li2019quality}, contrastive learning with proxy tasks~\cite{chen2021contrastive,madhusudana2023conviqt}, distortion-aware reconstruction via encoder-decoder architectures~\cite{xie2024qpt}, and learning-to-rank frameworks using pseudo-labeled video pairs~\cite{cao2025breaking}. Although these methods improve generalization to some extent by increasing the diversity of training data, they typically incur substantial costs in constructing large-scale datasets and demand significantly higher training overhead. In this paper, we demonstrate that VQAThinker enhances generalization without increasing the number of training samples, highlighting its efficiency.

\subsection{Explainable Video Quality Assessment}
Explainable VQA aims to identify the underlying distortions that impact perceptual quality. Some studies focus on developing multi-dimensional VQA models~\cite{wu2023towards,duan2025finevq} that produce quality scores across several dimensions, enabling diagnostic insights into the specific distortions affecting video quality. Recently, with the rapid advancement of LMMs, several works have proposed instruction-based datasets for image and video quality understanding, such as Q-Bench~\cite{wu2023q}, Q-Bench-Video~\cite{zhang2025vq}, Q-Instruct~\cite{wu2024q}, VQA$^2$~\cite{jia2024vqa}, and OmniVQA-Chat~\cite{jia2025scaling}, etc. These methods fine-tune LMMs on such datasets to generate descriptive language explanations of visual quality. While effective in producing interpretive outputs, these approaches require large-scale annotated datasets with detailed quality descriptions and often lack a direct connection to quantitative quality scoring.

DeepSeek-R1~\cite{guo2025deepseek} has shown remarkable effectiveness in promoting reasoning and generalization capabilities of LLMs through reinforcement learning. Q-Insight~\cite{li2025q} leverages GRPO with a threshold-based binary reward for score prediction and a classification-based binary rewards for distortion type and severity level, to enable reasoning-driven IQA. VisualQuality-R1~\cite{wu2025visualquality} formulates IQA as a ranking problem and introduces a rank-based reinforcement learning framework with a continuous fidelity loss to model the relative quality between image pairs. As a concurrent work, VQ-Insight~\cite{zhang2025vq} extends Q-Insight by adopting a cold-start reinforcement learning strategy that first warms up LMMs with image-level scoring, then incorporates temporal modeling and multi-task rewards across three training stages to enhance VQA. In contrast to VQ-Insight, our method eliminates the need for a cold-start procedure and multi-stage training, and introduces the dedicated VQA-specific rewards to achieve superior performance and efficiency.

\begin{figure*}[t]
\centering
\centerline{\epsfig{figure=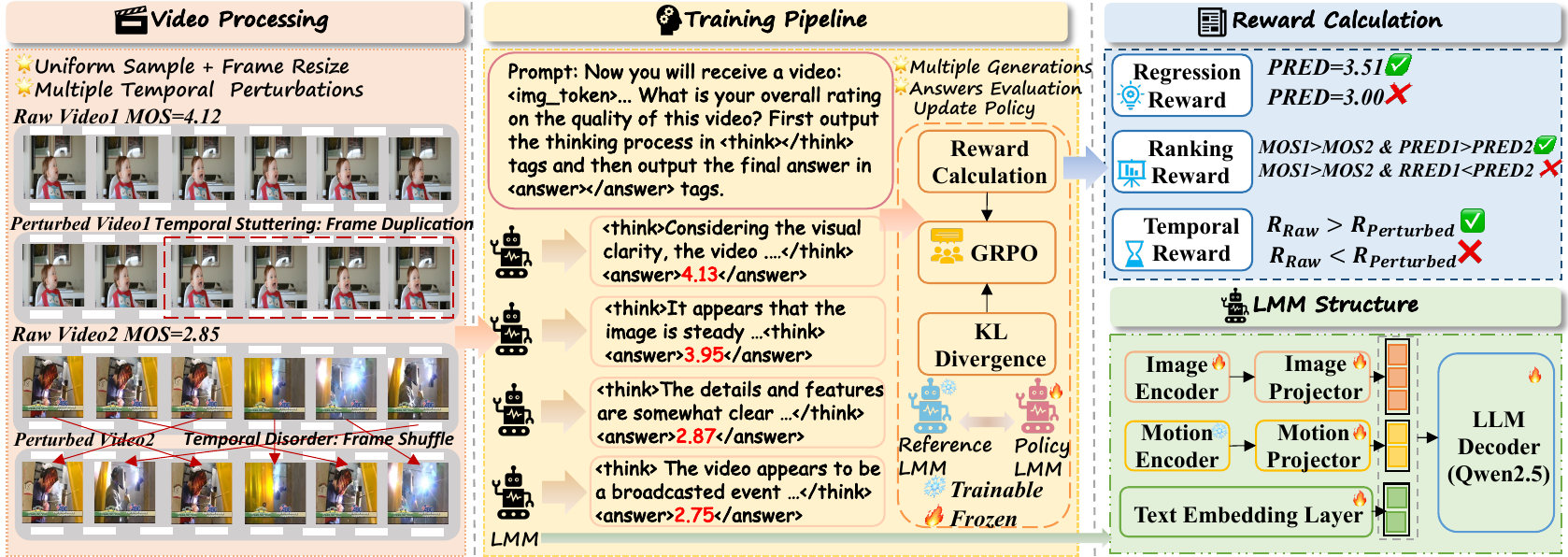,width=18cm}}
\caption{Overall framework of VQAThinker. The architecture consists of a LMM equipped with a frozen motion encoder and a motion projector. During training, the raw video is degraded using a frame perturbation operator to generate a perturbed version. Regression and ranking rewards are computed for both the raw and perturbed videos, while the temporal reward is obtained by comparing the reward differences between them. The regression, ranking, and temporal rewards derived from the raw videos are then used to optimize VQAThinker via the GRPO algorithm.}
\label{fig:model_structure}
\end{figure*}

\section{Method}
\label{sec:vqathinker}

\subsection{Model Architecture} 
\label{sec:model_architecture}
As illustrated in Figure~\ref{fig:model_structure}, we utilize an off-the-shelf LMM as the backbone of VQAThinker, which takes a video $\bm v$ and a text prompt $p$ as input, and produces a quality-aware response $q$ that consists of a quality reasoning trace $z$ and a scalar quality score $s$, delimited by \texttt{<think>}~\texttt{</think>} and \texttt{<answer>}~\texttt{</answer>} tags, respectively. To enable temporal-aware quality modeling, we additionally incorporate a frozen motion feature extractor to extract local temporal dynamics, along with a motion projector that maps the extracted temporal features into the language space. The detailed model architecture and video preprocessing procedure are provided in the supplementary material. We formalize this inference process as:
\begin{equation}
q = \mathcal{Q}_\theta(p, \bm{v}, \mathcal{M}(\bm{v})),
\end{equation}
where $\mathcal{Q}$ denotes the LMM augmented with the motion projector parameterized by fine-tuned weights $\theta$ and $\mathcal{M}$ represents the frozen motion feature extractor.

\subsection{Reinforcement Learning Strategy}

We adopt GRPO to train VQAThinker, which eliminates the need for an explicit value function by leveraging relative comparisons among grouped responses. During training, for a batch of videos $\mathcal{V} = \{\bm{v}_1, \bm{v}_2, ..., \bm{v}_N\}$, where $N$ denotes the batch size, GRPO generates $K$ quality-aware response for $\bm{v}_i$, denoted as $q(\bm{v}_i) = [ q_1(\bm{v}_i), q_2(\bm{v}_i), ..., q_K(\bm{v}_i)]$ using the previous policy $\pi_{\theta_{\text{old}}}$, parameterized by the weights from the previous training epoch $\theta_{\text{old}}$. These responses are then evaluated by \textbf{VQA-specific rewards} to obtain reward values $r(\bm{v}_i) = [ r_1(\bm{v}_i), r_2(\bm{v}_i), ..., r_K(\bm{v}_i)]$.

To evaluate the relative merit of the $k$-th response $r_k(\bm{v}_i)$ within the group of responses for video $\bm{v}_i$, GRPO computes its standardized advantage score $a_k(\bm{v}_i)$ by normalizing its reward over the $K$ responses:
\begin{equation}
\label{eq:advantage}
    a_k(\bm{v}_i) = \frac{r_k(\bm{v}_i) - \mu(r(\bm{v}_i))}{\sigma(r(\bm{v}_i))},
\end{equation}
where $\mu(\cdot)$ and $\sigma(\cdot)$ denote the mean and standard deviation operators respectively.

The optimization objective of GRPO is defined as:
\begin{align}
\mathcal{J}(\theta) =\ 
& \mathbb{E}_{\bm{v}_i \sim \mathcal{V},\ q_k \sim \pi_{\theta_{\text{old}}}(\cdot|\bm{v}_i)} \Big[
    \min\big(\rho_i a_k(\bm{v}_i),\ \bar{\rho}_i a_k(\bm{v}_i)\big) \nonumber\\
& \qquad -\ \beta \cdot \mathbb{D}_{\text{KL}}(\pi_\theta(q_k|\bm{v}_i) \,\|\, \pi_{\text{ref}}(\pi_\theta(q_k|\bm{v}_i))
\Big],
\end{align}
where $\rho_i = \frac{\pi_{\theta}(q_k \mid \bm{v}_i)}{\pi_{\theta_{\text{old}}}(q_k \mid \bm{v}_i)}$ is the importance sampling ratio between the current policy $\pi_{\theta}$ and the previous policy $\pi_{\theta_{\text{old}}}$, adjusting the magnitude of the advantage-weighted reward. $\bar{\rho}_i = \text{clip}(\rho_i,\ 1{-}\varepsilon_s,\ 1{+}\varepsilon_s)$ clips the ratio within a small range controlled by $\varepsilon_s > 0$ to ensure stable policy updates. $\beta$ is a coefficient that balances the KL divergence regularization term, which penalizes deviation from a reference policy $\pi_{\text{ref}}$, typically instantiated by the model parameters before reinforcement learning.

\begin{table*}[t]
    \centering
    {\scriptsize
    \setlength{\tabcolsep}{3pt}
    \renewcommand{\arraystretch}{0.8}
    \begin{tabular}{l l l cc cc cc cc cc cc}
        \toprule
        \multicolumn{3}{c}{\textbf{In-domain Datasets}} & \multicolumn{2}{c}{\textbf{LSVQ$_\text{test}$}} & \multicolumn{2}{c}{\textbf{LSVQ$_\text{1080p}$}} &\multicolumn{2}{c}{\textbf{KoNViD-1k}} & \multicolumn{2}{c}{\textbf{LIVE-VQC}} & \multicolumn{2}{c}{\textbf{YouTube-UGC}} & \multicolumn{2}{c}{\textbf{Overall}}\\
        \cmidrule(lr){1-3} \cmidrule(lr){4-5} \cmidrule(lr){6-7} \cmidrule(lr){8-9} \cmidrule(lr){10-11} \cmidrule(lr){12-13} \cmidrule(lr){14-15} 
        \multicolumn{3}{c}{\textbf{\# of videos}} & \multicolumn{2}{c}{\textbf{7,182}} & \multicolumn{2}{c}{\textbf{3,573}} & \multicolumn{2}{c}{\textbf{1,200}} & \multicolumn{2}{c}{\textbf{585}} & \multicolumn{2}{c}{\textbf{1,020}} & \multicolumn{2}{c}{\textbf{---}} \\
        \cmidrule(lr){1-3} \cmidrule(lr){4-5} \cmidrule(lr){6-7} \cmidrule(lr){8-9} \cmidrule(lr){10-11} \cmidrule(lr){12-13} \cmidrule(lr){14-15} 
        \textbf{Methods} & \textbf{Publication} & \textbf{Training data}  & \textbf{SRCC} & \textbf{PLCC} & \textbf{SRCC} & \textbf{PLCC} & \textbf{SRCC} & \textbf{PLCC} & \textbf{SRCC} & \textbf{PLCC} & \textbf{SRCC} & \textbf{PLCC}  & \textbf{SRCC} & \textbf{PLCC} \\
        \hline
        \rowcolor{gray!20}
\multicolumn{15}{l}{\textit{Unsupervised Methods}} \\
        NIQE & SPL'13 & \multirow{4}{*}{None} & 0.442 & 0.332 & 0.489 & 0.459 & 0.541 & 0.553 & 0.596 & 0.628 & 0.278 & 0.290 & 0.457 & 0.395 \\ 
        VIIDEO & TIP'15 & & 0.080 & 0.080 & 0.009 & 0.019  & 0.299 & 0.300 & 0.033 & 0.215 & 0.058 & 0.154 & 0.077 & 0.095 \\ 
        STEM & TIP'21 & & 0.206 & 0.243 & 0.434 & 0.381 & 0.619 & 0.627 & 0.594 & 0.629 & 0.284  & 0.318 & 0.325 & 0.336 \\
        CLIP-IQA& AAAI'23&  & 0.438 & 0.413 & 0.553 & 0.505 &  0.696 & 0.651 & 0.704 & 0.683 & 0.391 & 0.394 & 0.499 & 0.469 \\
        \hdashline
        \rowcolor{gray!20}
\multicolumn{15}{l}{\textit{Supervised Methods}} \\        
        SimpleVQA & MM'22& LSVQ (28K) & 0.864 & 0.861 & 0.756 & 0.801 & 0.861 & 0.860 & 0.762 & 0.799 & 0.808 & 0.808 & 0.827 & 0.838 \\
        FAST-VQA & ECCV'22& LSVQ (28K) & 0.880 & 0.880 & 0.781 & 0.813 & 0.859 & 0.854 & \textbf{0.826} & \underline{0.845} & 0.730 & 0.747 & 0.838 & 0.849  \\
        DOVER & ICCV'23& LSVQ (28K) & 0.878 & 0.866 & 0.782 & 0.813 & 0.874 & 0.869 & \underline{0.817} & 0.840 & 0.771 & 0.781 & 0.842 & 0.845  \\
        MinimalisticVQA &TPAMI'24 & LSVQ (28K) & \underline{0.885} & \underline{0.882} & 0.792 & \underline{0.828} & 0.862 & 0.859 & 0.775 & 0.821 & 0.826 & 0.821 & \underline{0.849} & 0.859 \\
        Q-Align & ICML'24& fused (287K) & \textbf{0.886} & \textbf{0.884} & 0.761 & 0.822 & \underline{0.876} & 0.878 & 0.783 & 0.819 & \underline{0.834} & \underline{0.846} & 0.844 & \underline{0.861} \\
        VQA$^2$-Scorer & MM'25 & VQA$^2$-ID (157K) & 0.878 & 0.872 & \underline{0.794} & 0.821 & \textbf{0.881} & \underline{0.880} & 0.785 & 0.830 & 0.811 & 0.823 & 0.847 & 0.854 \\
        \hdashline
        \rowcolor{gray!20}
\multicolumn{15}{l}{\textit{RL-based Methods}} \\        
        Q-Insight& NeurIPS'25 & LSVQ (28K) & 0.644 & 0.639 & 0.601 & 0.648 & 0.751 & 0.753 & 0.624 & 0.708 & 0.560 & 0.591 & 0.635 & 0.651 \\
        VisualQuality-R1& NeurIPS'25 & LSVQ (28K) & 0.795 & 0.796 & 0.716 & 0.744 & 0.784 & 0.792 & 0.732 & 0.781 & 0.717 & 0.730 & 0.765 & 0.776 \\
        VQ-Insight & arXiv'25 & LSVQ \& KonIQ (38K) & 0.875 & 0.876 & 0.786 & 0.823 & 0.875 & \textbf{0.884} & 0.790 & 0.835 & NA & NA & NA & NA \\
         \textbf{VQAThinker} & --- & LSVQ (28K) & 0.883 & 0.880 & \textbf{0.798} & \textbf{0.834} & \textbf{0.881} & \textbf{0.884} & 0.808 & \textbf{0.847} & \textbf{0.860} & \textbf{0.863} & \textbf{0.855} & \textbf{0.866} \\
        \midrule
        \multicolumn{3}{c}{\textbf{Out-of-distribution Datasets}} & \multicolumn{2}{c}{\textbf{LIVE-YT-Gaming}} & \multicolumn{2}{c}{\textbf{CGVDS}} & \multicolumn{2}{c}{\textbf{LIVE-YT-HFR}} & \multicolumn{2}{c}{\textbf{Waterloo-IVC-4K}} & \multicolumn{2}{c}{\textbf{VDPVE}} & \multicolumn{2}{c}{\textbf{Overall}} \\
        \cmidrule(lr){1-3} \cmidrule(lr){4-5} \cmidrule(lr){6-7} \cmidrule(lr){8-9} \cmidrule(lr){10-11} \cmidrule(lr){12-13} \cmidrule(lr){14-15} 
        \multicolumn{3}{c}{\textbf{\# of videos}} & \multicolumn{2}{c}{\textbf{600}} & \multicolumn{2}{c}{\textbf{357}} & \multicolumn{2}{c}{\textbf{480}} & \multicolumn{2}{c}{\textbf{1,200}} & \multicolumn{2}{c}{\textbf{839}} & \multicolumn{2}{c}{\textbf{---}} \\
        \cmidrule(lr){1-3} \cmidrule(lr){4-5} \cmidrule(lr){6-7} \cmidrule(lr){8-9} \cmidrule(lr){10-11} \cmidrule(lr){12-13} \cmidrule(lr){14-15}  
        \textbf{Methods} & \textbf{Publication}& \textbf{Training data} & \textbf{SRCC} & \textbf{PLCC} & \textbf{SRCC} & \textbf{PLCC} & \textbf{SRCC} & \textbf{PLCC} & \textbf{SRCC} & \textbf{PLCC} & \textbf{SRCC} & \textbf{PLCC}  & \textbf{SRCC} & \textbf{PLCC} \\
        \hline
        \rowcolor{gray!20}
\multicolumn{15}{l}{\textit{Unsupervised Methods}} \\
        NIQE & SPL'13 & \multirow{4}{*}{None} & 0.240 & 0.247 & 0.473 & 0.496 & 0.354 & 0.413 & 0.048 & 0.002 & 0.415 & 0.335 & 0.256 & 0.232 \\ 
        VIIDEO & TIP'15 & & 0.077 & -0.199 & 0.157 & -0.257 & 0.276 & 0.244 & 0.114 & 0.078 & 0.298 & 0.254 &  0.179 & 0.061 \\ 
        STEM & TIP'21& & 0.103 & 0.111 & 0.498 & 0.492 & 0.288 & 0.317 & 0.184 & 0.097 & 0.389 & 0.313 & 0.266 & 0.223\\ 
        CLIP-IQA& AAAI'23&  & 0.358 & 0.384 & 0.556 & 0.565 & 0.362 & 0.355 & 0.133 & 0.213 & 0.415 & 0.446 & 0.269 & 0.281\\
        \hdashline
        \rowcolor{gray!20}
\multicolumn{15}{l}{\textit{Supervised Methods}} \\        
        SimpleVQA &  MM'22 & LSVQ (28K) & 0.657 & 0.728 & 0.773 & 0.809 & \underline{0.416} & \underline{0.502} & 0.379 & 0.425 & 0.643 & 0.647  & 0.536 & 0.581\\
        FAST-VQA& ECCV'22 & LSVQ (28K) & 0.631 & 0.677 & 0.725 & 0.747 & 0.326 & 0.415 & 0.327 & 0.363 & 0.611 & 0.620 & 0.489 & 0.526 \\
        DOVER & ICCV'23 & LSVQ (28K) & 0.647 & 0.728 & 0.694 & 0.747 & 0.360 & 0.465 & 0.368 & 0.418 & 0.627 & 0.631 & 0.511 & 0.563\\
        MinimalisticVQA & TPAMI'24 & LSVQ (28K) & \underline{0.686} & \underline{0.746} & \underline{0.797} & \underline{0.816} & 0.301 & 0.388 & \underline{0.459} & \underline{0.502} & 0.639 & 0.641 & \underline{0.555} & \underline{0.594} \\
        Q-Align & ICML'24 & fused (287K) & 0.611 & 0.681 & 0.756 & 0.798  & 0.329 & 0.342 & 0.414 & 0.497 & 0.639 & 0.649 & 0.526 & 0.575 \\
        VQA$^2$-Scorer& MM'25 & VQA$^2$-ID (157K) & 0.613 & 0.698 & 0.656 & 0.741 & 0.332 & 0.413 & 0.415 & 0.474 & \underline{0.684} & \underline{0.692} & 0.527 & 0.584 \\
        \hdashline
        \rowcolor{gray!20}
\multicolumn{15}{l}{\textit{RL-based Methods}} \\        
        Q-Insight & NeurIPS'25 & LSVQ (28K) & 0.310 & 0.326 & 0.372 & 0.384 & 0.256 & 0.268 & 0.218 & 0.206 & 0.547 & 0.564 & 0.334 & 0.340 \\
        VisualQuality-R1& NeurIPS'25 & LSVQ (28K) & 0.472 & 0.548 & 0.493 & 0.574 & 0.340 & 0.347 & 0.227 & 0.298 & 0.622 & 0.637 & 0.408 & 0.458 \\
        VQ-Insight& arXiv'25 & LSVQ \& KonIQ (38K) & NA & NA & NA & NA & NA & NA & NA & NA & NA & NA & NA & NA  \\
         \textbf{VQAThinker} & --- & LSVQ (28K) & \textbf{0.767} & \textbf{0.806} & \textbf{0.856} & \textbf{0.845} & \textbf{0.528} & \textbf{0.610} & \textbf{0.573} & \textbf{0.624} & \textbf{0.706} & \textbf{0.716} & \textbf{0.661} & \textbf{0.698} \\
        \bottomrule
        
    \end{tabular}
    }
    \caption{Performance comparison of our model against other competitive methods. The best results are highlight in \textbf{boldface}, the second best is \underline{underlined}. NA in the table indicates unavailable results. ``Overall" represents the weighted average results based on the number of videos in each dataset.}
    \label{tab:performance}
\end{table*}

\subsection{VQA-specific Rewards}

The reward function provides training signals that guide the policy toward desirable behaviors. To this end, we design a set of VQA-specific reward functions that enable the model to learn video quality understanding and quality scoring. For the response $q_k^{i}$ (a simplified notation for $q_k(\bm{v}_i)$), its rewards include:

\noindent\textbf{Format Reward.}
To ensure that the model response $q_k^{i}$ conforms to the expected format, where the reasoning trace is enclosed within \texttt{<think>} tags and the predicted score is enclosed within \texttt{<answer>} tags, we introduce a binary format reward inspired by DeepSeek-R1, defined as:
\begin{equation}
\mathcal{R}_{\text{fmt}}(q_k^{i}) = \mathbb{I}\left[ \text{PatternMatch}(q_k^{i}) \right],
\end{equation}
where $\mathbb{I}[\cdot]$ denotes the indicator function that equals 1 when the condition holds and 0 otherwise, and $\text{PatternMatch}(q_k^{i})$ returns \texttt{True} if the output $q_k^{i}$ satisfies the required format constraints.

\noindent\textbf{Bell-shaped Regression Reward.} VQA is naturally formulated as a regression task. However, existing regression-based rewards struggle to provide fine-grained training signals. Some methods use constant binary rewards~\cite{li2025q}, assigning fixed values when predictions fall within a threshold, without capturing their proximity to the ground truth. Others adopt L1-norm-based rewards~\cite{zhang2025vq}, where the reward scales linearly with prediction error at a constant rate. In practice, learning becomes increasingly difficult as the predicted score approaches the ground truth. Therefore, the reward function should provide stronger and more rapidly changing signals in this fine-grained regime to better guide the model toward high-precision predictions.

To address this, we propose a bell-shaped regression reward based on a Gaussian function, which increases rapidly as the prediction error decreases and becomes progressively less sensitive near the ground truth, thereby facilitating fine-grained score optimization:

\begin{equation}
\mathcal{R}_{\text{reg}}(q_k^{i}) = \alpha \cdot \exp\left(-\frac{(s_k^{i} - g^{i})^2}{2\sigma^2}\right),
\label{eq:div}
\end{equation}
where $s_k^{i}$ is the predicted quality score extracted from the response $q_k^{i}$, $g^{i}$ is the ground-truth quality score of $\bm{v}_i$, $\sigma$ controls the sensitivity of the reward to deviations from the ground truth, and $\alpha \in (0, 1]$ is a scaling factor that modulates the overall magnitude of the reward.

\noindent\textbf{Pairwise Ranking Reward.} VQA can also be formulated as a ranking problem, where the objective is to learn the relative quality ordering between video pairs based on their predicted scores. This formulation is typically optimized using ranking-based loss functions, such as fidelity loss~\cite{tsai2007frank} or approximate Spearman’s Rank Correlation Coefficient (SRCC) loss~\cite{li2022blindly}, which measure the consistency between predicted and ground-truth rankings. 

Inspired by VisualQuality-R1~\cite{wu2025visualquality}, we adopt a pairwise ranking reward that assigns higher rewards to video pairs whose predicted ranking aligns with the ground-truth rank order. Specifically, for $s_k^{i}$, the predicted score for the $k$-th response $q_k^{i}$ of video $\bm{v}_i$, we compare it against the average prediction $\mu({s}_j)$ of another video $\bm{v}_j$ to evaluate whether the predicted ranking aligns with the ground-truth rank order between $g_i$ and $g_j$. To quantify the likelihood that $s_k^{i}$ ranks higher than $\mu({s}_j)$, we compute comparative probability $p_k(\bm{v}_i,\bm{v}_j)$ using the cumulative distribution function of the standard normal distribution:
\[
p_k(\bm{v}_i,\bm{v}_j) = \Phi\left(\frac{s_{i}^{j} - \mu({s}_j)}{\sqrt{\sigma(s_i)^2 + \sigma(s_j)^2 + \epsilon}}\right),
\]
where $\sigma(s_i)^2$ and $\sigma(s_j)^2$ denote the variance of $K$ predicted scores for $\bm{v}_i$ and $\bm{v}_j$, respectively, and $\epsilon$ is a small positive constant for numerical stability, and the subsequent occurrences $\epsilon$ serve the same purpose. 

The pairwise ranking reward is computed based on the fidelity loss~\cite{tsai2007frank}, and is defined as:
\begin{align}
\mathcal{R}_{\text{rank}}(q_k^{i}) &= \sqrt{p_k(\bm{v}_i,\bm{v}_j) \cdot \mathbb{I}[g_i > g_j] + \epsilon} \\
& \quad+ \sqrt{(1 - p_k(\bm{v}_i,\bm{v}_j)) \cdot \mathbb{I}[g_i < g_k] + \epsilon},
\end{align}
where $\mathbb{I}[\cdot]$ is the indicator function.

\noindent\textbf{Temporal Consistency Reward.} 
We have introduced a motion feature extractor to capture local temporal dynamics. However, this module is limited in handling long-range temporal distortions that may occur across the sampled frames. 

To address this issue, we propose a temporal consistency reward that explicitly encourages the model to be sensitive to broader temporal degradations. Specifically, for video $\bm{v}_i$, we construct a temporally perturbed counterpart $\bm{v}_{i,\text{temp}}$ by applying various frame-level distortions, such as frame repetition and frame dropping, to simulate real-world temporal artifacts like frame lagging and frame loss (see supplementary material for details). For the responses $q_i$ and $q_{i,\text{temp}}$, we compute both the regression reward $\mathcal{R}_{\text{reg}}(\cdot)$ and the ranking reward $\mathcal{R}_{\text{rank}}(\cdot)$. The temporal consistency reward is then defined to encourage model behavior that yields a higher regression or ranking reward for $\bm{v}_i$ than for $\bm{v}_{i,\text{temp}}$:

\begin{equation}
\mathcal{R}_{\text{temp}}^{(t)} (q_k^i) =
\begin{cases}
\delta, & \text{if } \mu(r_{i}^{(t)}) \geq \mu(r_{i,\text{temp}}^{(t)})) \text{ and } \mu(r_{i}^{(t)}) > \tau \\
0, & \text{otherwise}
\end{cases}
\label{eq:temp_subreward}
\end{equation}
where $t \in \{\text{reg}, \text{rank}\}$ indicates the reward type, $\mu(\cdot)$ denotes the average reward across all responses, $\delta$ is a fixed bonus, and $\tau$ is a confidence threshold to ensure that temporal consistency rewards are only granted when the model is sufficiently accurate. The final temporal consistency reward~\footnote{Note that the rewards from $\bm{v}_{i,\text{temp}}$ are used only for comparison, and are not included in the reinforcement learning objective.} is the sum of the two sub-rewards:
\begin{equation}
\mathcal{R}_{\text{temp}} (q_k^i) = \mathcal{R}_{\text{temp}}^{(\text{reg})} (q_k^i) + \mathcal{R}_{\text{temp}}^{(\text{rank})} (q_k^i).
\end{equation}

The final reward is computed as the sum of the three components:
\begin{equation}
r_k(\bm{v}_i) = \mathcal{R}_{\text{fmt}}(q_k^i) + \mathcal{R}_{\text{reg}}(q_k^i) + \mathcal{R}_{\text{rank}}(q_k^i) + \mathcal{R}_{\text{temp}}(q_k^i).
\end{equation}

\section{Experiments}
\subsection{Experimental Setups}

\noindent\textbf{Benchmark Datasets.} 
VQAThinker is trained on the training set of LSVQ~\cite{ying2021patch} containing $28,056$ UGC videos. We evaluate VQAThinker from two perspectives: \textbf{video quality scoring} and \textbf{video quality understanding}. 

For \textbf{video quality scoring}, we evaluate performance on ten VQA benchmarks: LSVQ Test~\cite{ying2021patch}, LSVQ 1080p~\cite{ying2021patch}, KoNViD-1k~\cite{hosu2017konstanz}, LIVE-VQC~\cite{sinno2018large}, YouTube-UGC~\cite{wang2019youtube}, LIVE-YT-Gaming~\cite{yu2022subjective}, CGVDS~\cite{saha2023study}, LIVE-YT-HFR~\cite{madhusudana2021subjective}, Waterloo-IVC-4K~\cite{li2019avc}, and VDPVE~\cite{gao2023vdpve}. The first five datasets are considered in-domain, as they primarily consist of UGC videos with in-the-wild distortions. The remaining five datasets are treated as OOD due to their distinct content types (\textit{e.g.,} gaming videos) or distortion patterns (\textit{e.g.,} compression and enhancement artifacts). For \textbf{video quality understanding}, we adopt two evaluation benchmarks: FineVD~\cite{duan2025finevq} and Q-Bench-Video~\cite{zhang2025q}. FineVD is used to assess the model's ability to identify quality-related distortion attributes, while Q-Bench-Video measures the accuracy of video quality descriptions.


\begin{table}[t]
\centering
{\scriptsize
\setlength{\tabcolsep}{4.5pt}
\renewcommand{\arraystretch}{0.8}
\begin{tabular}{lcccccc}
\toprule
\textbf{Model / Attribute} & \textbf{Color} & \textbf{Noise} & \textbf{Artifact} & \textbf{Blur} & \textbf{Temporal} & \textbf{Overall} \\
\hline
\rowcolor{gray!20}
\multicolumn{7}{l}{\textit{Open-source LMMs}} \\
VideoChat2  & 18.25& 18.80& 24.26& 31.15& 15.19 & 21.53 \\
VideoLLaMA2  & 57.87& 63.98& 51.87& 61.91& \textbf{86.42} & 64.41 \\
Video-ChatGPT & 28.35& 27.66& 44.49& 62.89& 18.60 & 36.40 \\
MiniCPM-V  & 70.47& 65.85& \underline{52.66} & 46.26& 71.95 & 61.44 \\
LLaVA-NeXT & 26.87& 27.26& 43.90& 64.76& 13.09 & 35.18 \\
Video-LLaVA  & 26.48& 27.26& 43.90& 64.67& 13.09 & 35.08 \\
Qwen2.5-VL & 68.49 & 48.74 & 47.51 & 52.08 & 56.25 & 54.61  \\
InternVL3 & 64.63& 62.18& 52.49& 52.98& 68.01 &60.06 \\
\hdashline
\rowcolor{gray!20}
\multicolumn{7}{l}{\textit{Explainable VQAs}} \\
FineVQ & \underline{73.52} & \textbf{72.74} & 51.87& 64.76& 65.06 & \underline{65.59} \\
Q-Insight   & 63.99& 52.52& 48.21& 64.58& 68.01 & 59.46 \\
VisualQuality-R1  & 65.27& 55.46& 48.17& \underline{65.77} & 63.24 & 59.58 \\
\rowcolor{gray!10}
\textbf{VQAThinker} & \textbf{82.64} & \underline{70.17} & \textbf{55.15} & \textbf{70.83} & \underline{83.46} & \textbf{72.45} \\
\bottomrule
\end{tabular}}
\caption{Performance comparison of explainable VQA methods and LMMs on the FineVD dataset for the video distortion attribution task. All values are in \% (percentage).}
\label{tab:finevd}
\end{table}

\noindent\textbf{Competing Methods} We compare VQAThinker with three categories of methods: 1) unsupervised methods: NIQE~\cite{mittal2012making}, VIIDEO~\cite{mittal2015completely}, STEM~\cite{kancharla2021completely}, and CLIP-IQA~\cite{wang2023exploring}; 2) supervised methods: SimpleVQA~\cite{sun2022deep}, FAST-VQA~\cite{wu2022fast}, DOVER~\cite{wu2023exploring}, MinimalisticVQA (IX)~\cite{sun2024analysis}, Q-Align~\cite{wu2023qalign} and VQA$^2$-Scorer~\cite{jia2024vqa}; 3) reinforcement learning methods: Q-Insight~\cite{li2025q}, VisualQuality-R1~\cite{wu2025visualquality}, and VQ-Insight~\cite{zhang2025vq}. To ensure fairness, we retrain the IQA-based methods such as Q-Insight and VisualQuality-R1 on the LSVQ training set, using the same sampled frames as VQAThinker to adapt them for the VQA task.\footnote{Since VQ-Insight is not open-sourced, we report its performance based on the official results.}

\noindent\textbf{Implementation Details.} 
We utilize InternVL3-8B~\cite{zhu2025internvl3} as the backbone LMM and SlowFast~\cite{feichtenhofer2019slowfast} as the motion feature extractor, both initialized with their publicly released weights. During training, the motion feature extractor is kept frozen, while all other components are fine-tuned. In the GRPO trainer, the number of response $K$ is set to 4, and the weight of the KL penalty $\beta$ is set to 0.04. The model is trained for three epochs on 8 NVIDIA A800 GPUs with a total batch size $N = 64$ and a learning rate of $1 \times 10^{-6}$. We set $\sigma = 0.5$ and $\alpha = 0.8$ in Eq.~\eqref{eq:div}, and $\delta = 0.3$ and $\tau = 0.5$ in Eq.~\eqref{eq:temp_subreward}.

\subsection{Performance Analysis}
\noindent\textbf{Video Quality Scoring.} 
The SRCC and PLCC results across ten VQA benchmarks are summarized in Table~\ref{tab:performance}. For in-domain evaluation, VQAThinker achieves state-of-the-art performance on nearly all benchmarks and outperforms all competing methods in terms of overall accuracy. Compared to VQ-Insight, a recent RL-based VQA method, VQAThinker demonstrates consistently superior performance on the available benchmarks, despite adopting a more streamlined training process with fewer training samples. For OOD evaluation, VQAThinker exhibits a substantial advantage over existing methods, regardless of content domain (\textit{e.g.,} gaming videos in LIVE-YT-Gaming and CGVDS) or distortion type (\textit{e.g.,} frame rate in LIVE-YT-HFR, compression in Waterloo-IVC-4K, and enhancement in VDPVE). Overall, it achieves a $19.1\%$ relative improvement in average SRCC on OOD datasets, without relying on additional training samples to increase data diversity. These results demonstrate the effectiveness and strong generalization capability of our approach.

\begin{table}[t]
\centering
{\scriptsize
\setlength{\tabcolsep}{12pt}
\renewcommand{\arraystretch}{0.8}
\begin{tabular}{lccc}
\toprule
\textbf{Model} & \textbf{Global}↑ & \textbf{Referring}↑ & \textbf{Overall}↑ \\
\hline
\rowcolor{gray!20}
\multicolumn{4}{l}{\textit{Open-source LMMs}} \\
VideoChat2   & 38.91& 40.47& 39.68 \\
VILA1.5   & 46.55& 48.69& 47.61 \\
InternVL-Chat   & 49.43& 54.05& 51.72 \\
LLaVA-OneVision   & 49.41& 49.35& 49.38 \\
LLaVA-Next-Video & 47.01& 52.23& 49.59 \\
mPLUG-Owl3  & 51.71& 48.78& 50.26 \\
Video-LLaVA  & 43.66& 47.73& 45.67 \\
Qwen2.5-VL & 55.71 & 53.75 & 54.66 \\
InternVL3 & 50.87& 50.75& 50.80\\
\hdashline
\rowcolor{gray!20}
\multicolumn{4}{l}{\textit{Proprietary LMMs}} \\
Gemini 1.5 Pro & 51.98& \textbf{60.80} & 56.35 \\
GPT-4o & 57.98& 54.39& 56.17 \\
GPT-4o mini & 50.03& 49.88& 49.95 \\
\hdashline
\rowcolor{gray!20}
\multicolumn{4}{l}{\textit{Explainable VQAs}} \\
VQA$^2$-Assistant   & \textbf{61.59} & \underline{57.36} & \textbf{59.32} \\
Q-Insight   & 55.71& 53.45& 54.50 \\
VisualQuality-R1  & 57.09& 54.35& 55.63 \\
\rowcolor{gray!10}
\textbf{VQAThinker} & \underline{58.82} & 56.46& \underline{57.56} \\

\bottomrule
\end{tabular}}
\caption{Performance comparison of explainable VQA methods and LMMs on the Q-Bench-Video (dev) dataset for the video quality description task. All values are in \%.}
\label{tab:q-bench}
\end{table}

\noindent\textbf{Video Quality Understanding.}
The results on FineVD and Q-Bench-Video are presented in Table~\ref{tab:finevd} and Table~\ref{tab:q-bench} respectively. \textit{Notably, VQAThinker is not explicitly trained on any video quality instruction datasets; instead, we directly evaluate its reasoning traces as a measure of the model’s quality understanding ability.} Despite this, VQAThinker achieves strong performance on both benchmarks. On FineVD, which assesses distortion attribution capabilities, VQAThinker significantly outperforms open-source LMMs and RL-based quality assessment models. Remarkably, it even surpasses FineVQ, a model specifically fine-tuned on the FineVD training set. Similarly, on Q-Bench-Video, which evaluates the overall quality description ability, VQAThinker outperforms all open-source and proprietary LMMs as well as RL-based baselines, and is only $1.76\%$ below VQA$^2$-Assistant, a model trained on $157$K instruction–answer pairs. These results demonstrate that the proposed reinforcement learning framework, equipped with VQA-specific rewards, effectively enables the model to reason about potential video distortions that affect perceptual quality.

\begin{table}[t]
    
    \centering
    {\scriptsize
    \setlength{\tabcolsep}{3pt}
    \renewcommand{\arraystretch}{0.8}
    \begin{tabular}{ccc|c|c|cc|cc}
        \toprule
        \multicolumn{3}{c|} {\textbf{Reward}} & \multirow{2}{*}{\textbf{Motion}} & \multirow{2}{*}{\textbf{Reasoning}} &  \multicolumn{2}{c|}{\textbf{In-domain}} &  \multicolumn{2}{c}{\textbf{OOD}} \\
        \cmidrule(lr){1-3} 
        \cmidrule(lr){6-7} \cmidrule(lr){8-9} 
         \textbf{Rank} & \textbf{Reg.} & \textbf{Temp.} &  & &
        \textbf{SRCC} & \textbf{PLCC} & \textbf{SRCC} & \textbf{PLCC} \\
        \midrule

        \ding{52} &           &           &  \ding{52} & \ding{52} & 0.780 & 0.785 &  0.506 & 0.541 \\
                  & \ding{52} & &     \ding{52}      & \ding{52} &   0.823    &   0.834 &  0.620 & 0.652 \\
        \ding{52} & \ding{52} & & \ding{52} & \ding{52} &  0.849 & 0.860 &  0.638 & 0.677 \\
        \ding{52} &  \ding{52} &   &           & \ding{52} & 0.840  & 0.853 &  0.614 & 0.650 \\
        \ding{52} & \ding{52} & \ding{52} & \ding{52} &           & 0.844 & 0.848 &  0.627 & 0.660 \\
        \ding{52} &  \ding{52} &\ding{52} & \ding{52} & \ding{52} &  0.855 & 0.866 & 0.662 & 0.698 \\

        \bottomrule
    \end{tabular}
    }
    \caption{Performance of ablation studies. ``Rank", ``Reg.", and ``Temp." denote the pairwise ranking reward, bell-shaped regression reward, and temporal consistency reward, respectively. ``Motion" refers to whether the motion feature extractor are included, and ``Reasoning" indicates whether the reasoning output is required. The in-domain and OOD results are averaged across the corresponding test sets.}
    \label{tab:ablation}
\end{table}

\subsection{Ablation Study}

\noindent\textbf{Motion Feature Extractor.}
We incorporate a frozen motion feature extractor to capture local temporal dynamics. To assess its standalone effect, we ablate it under the setting without the temporal consistency reward. As shown in Table~\ref{tab:ablation}, introducing the motion feature extractor consistently improves both in-domain and OOD performance, validating its effectiveness in modeling local temporal distortions.

\noindent\textbf{VQA-specific Rewards.}
We first evaluate the individual impact of the bell-shaped regression reward and the pairwise ranking reward. From Table~\ref{tab:ablation}, the bell-shaped regression reward yields significantly better performance than the pairwise ranking reward, demonstrating its effectiveness in promoting fine-grained quality assessment. When the two rewards are combined, performance further improves, indicating that the pairwise ranking reward complements the regression objective by enhancing the model’s ability to discern the relative quality order between video pairs. Adding the temporal consistency reward, which is designed to account for long-term temporal distortions, leads to additional performance gains, validating its contribution to improving temporal sensitivity in VQA.

\noindent\textbf{Reasoning.}  
We incorporate the format reward to guide the model to generate a reasoning trace prior to producing the final quality score, which better aligns with the human perceptual decision-making process. To assess its impact, we remove the \texttt{\textless think\textgreater} tags from the format reward, disabling the reasoning generation. As shown in Table~\ref{tab:ablation}, we observe a consistent performance drop across both in-domain and OOD benchmarks, with more pronounced degradation on the OOD datasets. It indicates that the explicit reasoning procedure is important for enhancing the model’s video quality scoring capability.

\section{Conclusion}

In this work, we propose VQAThinker, a reasoning-based VQA framework that leverages GRPO to achieve generalizable and explainable VQA. Specifically, we introduce three VQA-specific rewards: a bell-shaped regression reward for fine-grained score prediction, a pairwise ranking reward for relative quality discrimination, and a temporal consistency reward for modeling temporal degradations. Extensive experiments across ten VQA benchmarks and two quality description datasets demonstrate that VQAThinker achieves state-of-the-art performance in both in-domain and OOD settings, while also providing strong interpretability through explicit reasoning traces. 


\section{Acknowledgements}
This work was supported in part by the National Natural Science Foundation of China under Grant 62522116, Grant 62271312, Grant 62132006, and Grant 62301316, and in part by STCSM under Grant 22DZ2229005.

\bibliography{aaai2026}

@String(AAAI = {AAAI})

@inproceedings{ying2021patch,
  title={Patch-vq:'patching up'the video quality problem},
  author={Ying, Zhenqiang and Mandal, Maniratnam and Ghadiyaram, Deepti and Bovik, Alan},
  booktitle={Proceedings of the IEEE/CVF Conference on Computer Vision and Pattern Recognition},
  pages={14019--14029},
  year={2021}
}

@inproceedings{wu2023exploring,
  title={Exploring video quality assessment on user generated contents from aesthetic and technical perspectives},
  author={Wu, Haoning and Zhang, Erli and Liao, Liang and Chen, Chaofeng and Hou, Jingwen and Wang, Annan and Sun, Wenxiu and Yan, Qiong and Lin, Weisi},
  booktitle={Proceedings of the IEEE/CVF International Conference on Computer Vision},
  pages={20144--20154},
  year={2023}
}

@inproceedings{wu2022fast,
  title={Fast-vqa: Efficient end-to-end video quality assessment with fragment sampling},
  author={Wu, Haoning and Chen, Chaofeng and Hou, Jingwen and Liao, Liang and Wang, Annan and Sun, Wenxiu and Yan, Qiong and Lin, Weisi},
  booktitle={European Conference on Computer Vision},
  pages={538--554},
  year={2022},
  organization={Springer}
}

@article{wu2023qalign,
  title={Q-align: Teaching lmms for visual scoring via discrete text-defined levels},
  author={Wu, Haoning and Zhang, Zicheng and Zhang, Weixia and Chen, Chaofeng and Liao, Liang and Li, Chunyi and Gao, Yixuan and Wang, Annan and Zhang, Erli and Sun, Wenxiu and others},
  journal={arXiv preprint arXiv:2312.17090},
  year={2023}
}

@article{sun2024analysis,
  title={Analysis of video quality datasets via design of minimalistic video quality models},
  author={Sun, Wei and Wen, Wen and Min, Xiongkuo and Lan, Long and Zhai, Guangtao and Ma, Kede},
  journal={IEEE Transactions on Pattern Analysis and Machine Intelligence},
  year={2024},
  publisher={IEEE}
}

@inproceedings{feichtenhofer2019slowfast,
  title={Slowfast networks for video recognition},
  author={Feichtenhofer, Christoph and Fan, Haoqi and Malik, Jitendra and He, Kaiming},
  booktitle={Proceedings of the IEEE/CVF International Conference on Computer Vision},
  pages={6202--6211},
  year={2019}
}

@incollection{thurstone2017law,
  title={A law of comparative judgment},
  author={Thurstone, Louis L},
  booktitle={Scaling},
  pages={81--92},
  year={2017},
  publisher={Routledge}
}

@inproceedings{hosu2017konstanz,
  title={The konstanz natural video database (KoNViD-1k)},
  author={Hosu, Vlad and Hahn, Franz and Jenadeleh, Mohsen and Lin, Hanhe and Men, Hui and Szir{\'a}nyi, Tam{\'a}s and Li, Shujun and Saupe, Dietmar},
  booktitle={International Conference on Quality of Multimedia Experience},
  pages={1--6},
  year={2017}
}

@inproceedings{gao2023vdpve,
  title={Vdpve: Vqa dataset for perceptual video enhancement},
  author={Gao, Yixuan and Cao, Yuqin and Kou, Tengchuan and Sun, Wei and Dong, Yunlong and Liu, Xiaohong and Min, Xiongkuo and Zhai, Guangtao},
  booktitle={Proceedings of the IEEE/CVF Conference on Computer Vision and Pattern Recognition},
  pages={1474--1483},
  year={2023}
}

@article{sinno2018large,
  title={Large-scale study of perceptual video quality},
  author={Sinno, Zeina and Bovik, Alan Conrad},
  journal={IEEE Transactions on Image Processing},
  volume={28},
  number={2},
  pages={612--627},
  year={2018},
  publisher={IEEE}
}

@inproceedings{wang2019youtube,
  title={YouTube UGC dataset for video compression research},
  author={Wang, Yilin and Inguva, Sasi and Adsumilli, Balu},
  booktitle={International Workshop on Multimedia Signal Processing},
  pages={1--5},
  year={2019},
  organization={IEEE}
}

@inproceedings{yu2022subjective,
  title={Subjective quality assessment of user-generated content gaming videos},
  author={Yu, Xiangxu and Tu, Zhengzhong and Ying, Zhenqiang and Bovik, Alan C and Birkbeck, Neil and Wang, Yilin and Adsumilli, Balu},
  booktitle={Proceedings of the IEEE/CVF Winter Conference on Applications of Computer Vision},
  pages={74--83},
  year={2022}
}

@article{li2024llava,
  title={Llava-onevision: Easy visual task transfer},
  author={Li, Bo and Zhang, Yuanhan and Guo, Dong and Zhang, Renrui and Li, Feng and Zhang, Hao and Zhang, Kaichen and Zhang, Peiyuan and Li, Yanwei and Liu, Ziwei and others},
  journal={arXiv preprint arXiv:2408.03326},
  year={2024}
}

@inproceedings{li2019avc,
  title={AVC, HEVC, VP9, AVS2 OR AV1?—A comparative study of state-of-the-art video encoders on 4K videos},
  author={Li, Zhuoran and Duanmu, Zhengfang and Liu, Wentao and Wang, Zhou},
  booktitle={Proceedings of the International Conference on Image Analysis and Recognition},
  pages={162--173},
  year={2019},
  organization={Springer}
}

@article{mittal2012making,
  title={Making a “completely blind” image quality analyzer},
  author={Mittal, Anish and Soundararajan, Rajiv and Bovik, Alan C},
  journal={IEEE Signal Processing Letters},
  volume={20},
  number={3},
  pages={209--212},
  year={2012},
  publisher={IEEE}
}

@article{ge2024lmm,
  title={LMM-VQA: Advancing video quality assessment with large multimodal models},
  author={Ge, Qihang and Sun, Wei and Zhang, Yu and Li, Yunhao and Ji, Zhongpeng and Sun, Fengyu and Jui, Shangling and Min, Xiongkuo and Zhai, Guangtao},
  journal={arXiv preprint arXiv:2408.14008},
  year={2024}
}

@article{saha2023study,
  title={Study of subjective and objective quality assessment of mobile cloud gaming videos},
  author={Saha, Avinab and Chen, Yu-Chih and Davis, Chase and Qiu, Bo and Wang, Xiaoming and Gowda, Rahul and Katsavounidis, Ioannis and Bovik, Alan C},
  journal={IEEE Transactions on Image Processing},
  volume={32},
  pages={3295--3310},
  year={2023},
  publisher={IEEE}
}

@article{madhusudana2021subjective,
  title={Subjective and objective quality assessment of high frame rate videos},
  author={Madhusudana, Pavan C and Yu, Xiangxu and Birkbeck, Neil and Wang, Yilin and Adsumilli, Balu and Bovik, Alan C},
  journal={IEEE Access},
  volume={9},
  pages={108069--108082},
  year={2021},
  publisher={IEEE}
}

@article{min2024perceptual,
  title={Perceptual video quality assessment: A survey},
  author={Min, Xiongkuo and Duan, Huiyu and Sun, Wei and Zhu, Yucheng and Zhai, Guangtao},
  journal={Science China Information Sciences},
  volume={67},
  number={11},
  pages={211301},
  year={2024},
  publisher={Springer}
}

@inproceedings{liu2022video,
  title={Video swin transformer},
  author={Liu, Ze and Ning, Jia and Cao, Yue and Wei, Yixuan and Zhang, Zheng and Lin, Stephen and Hu, Han},
  booktitle={Proceedings of the IEEE/CVF Conference on Computer Vision and Pattern Recognition},
  pages={3202--3211},
  year={2022}
}

@inproceedings{wu2024q,
  title={Q-instruct: Improving low-level visual abilities for multi-modality foundation models},
  author={Wu, Haoning and Zhang, Zicheng and Zhang, Erli and Chen, Chaofeng and Liao, Liang and Wang, Annan and Xu, Kaixin and Li, Chunyi and Hou, Jingwen and Zhai, Guangtao and others},
  booktitle={Proceedings of the IEEE/CVF Conference on Computer Vision and Pattern Recognition},
  pages={25490--25500},
  year={2024}
}

@article{kancharla2021completely,
  title={Completely blind quality assessment of user generated video content},
  author={Kancharla, Parimala and Channappayya, Sumohana S},
  journal={IEEE Transactions on Image Processing},
  volume={31},
  pages={263--274},
  year={2021},
  publisher={IEEE}
}

@article{mittal2015completely,
  title={A completely blind video integrity oracle},
  author={Mittal, Anish and Saad, Michele A and Bovik, Alan C},
  journal={IEEE Transactions on Image Processing},
  volume={25},
  number={1},
  pages={289--300},
  year={2015},
  publisher={IEEE}
}

@inproceedings{deng2009imagenet,
  title={Imagenet: A large-scale hierarchical image database},
  author={Deng, Jia and Dong, Wei and Socher, Richard and Li, Li-Jia and Li, Kai and Fei-Fei, Li},
  booktitle={IEEE Conference on Computer Vision and Pattern Recognition},
  pages={248--255},
  year={2009}
}

@inproceedings{carreira2017quo,
  title={Quo vadis, action recognition? a new model and the kinetics dataset},
  author={Carreira, Joao and Zisserman, Andrew},
  booktitle={proceedings of the IEEE Conference on Computer Vision and Pattern Recognition},
  pages={6299--6308},
  year={2017}
}

@article{chen2021contrastive,
  title={Contrastive self-supervised pre-training for video quality assessment},
  author={Chen, Pengfei and Li, Leida and Wu, Jinjian and Dong, Weisheng and Shi, Guangming},
  journal={IEEE Transactions on Image Processing},
  volume={31},
  pages={458--471},
  year={2021},
  publisher={IEEE}
}

@article{madhusudana2023conviqt,
  title={Conviqt: Contrastive video quality estimator},
  author={Madhusudana, Pavan C and Birkbeck, Neil and Wang, Yilin and Adsumilli, Balu and Bovik, Alan C},
  journal={IEEE Transactions on Image Processing},
  volume={32},
  pages={5138--5152},
  year={2023},
  publisher={IEEE}
}

@article{mittal2012no,
  title={No-reference image quality assessment in the spatial domain},
  author={Mittal, Anish and Moorthy, Anush Krishna and Bovik, Alan Conrad},
  journal={IEEE Transactions on Image Processing},
  volume={21},
  number={12},
  pages={4695--4708},
  year={2012},
  publisher={IEEE}
}

@article{konrad1992bayesian,
  title={Bayesian estimation of motion vector fields},
  author={Konrad, Janusz and Dubois, Eric},
  journal={IEEE Transactions on Pattern Analysis and Machine Intelligence},
  volume={14},
  number={09},
  pages={910--927},
  year={1992},
  publisher={IEEE Computer Society}
}

@article{beauchemin1995computation,
  title={The computation of optical flow},
  author={Beauchemin, Steven S. and Barron, John L.},
  journal={ACM Computing Surveys},
  volume={27},
  number={3},
  pages={433--466},
  year={1995}
}

@inproceedings{li2019quality,
  title={Quality assessment of in-the-wild videos},
  author={Li, Dingquan and Jiang, Tingting and Jiang, Ming},
  booktitle={Proceedings of the ACM international Conference on Multimedia},
  pages={2351--2359},
  year={2019}
}

@article{wu2023discovqa,
  title={Discovqa: Temporal distortion-content transformers for video quality assessment},
  author={Wu, Haoning and Chen, Chaofeng and Liao, Liang and Hou, Jingwen and Sun, Wenxiu and Yan, Qiong and Lin, Weisi},
  journal={IEEE Transactions on Circuits and Systems for Video Technology},
  volume={33},
  number={9},
  pages={4840--4854},
  year={2023},
  publisher={IEEE}
}

@inproceedings{sun2022deep,
  title={A deep learning based no-reference quality assessment model for ugc videos},
  author={Sun, Wei and Min, Xiongkuo and Lu, Wei and Zhai, Guangtao},
  booktitle={Proceedings of the ACM International Conference on Multimedia},
  pages={856--865},
  year={2022}
}

@inproceedings{xie2024qpt,
  title={QPT-V2: Masked image modeling advances visual scoring},
  author={Xie, Qizhi and Yuan, Kun and Qu, Yunpeng and Wu, Mingda and Sun, Ming and Zhou, Chao and Zhu, Jihong},
  booktitle={Proceedings of the ACM International Conference on Multimedia},
  pages={2709--2718},
  year={2024}
}

@article{li2025q,
  title={Q-insight: Understanding image quality via visual reinforcement learning},
  author={Li, Weiqi and Zhang, Xuanyu and Zhao, Shijie and Zhang, Yabin and Li, Junlin and Zhang, Li and Zhang, Jian},
  journal={arXiv preprint arXiv:2503.22679},
  year={2025}
}

@article{wu2025visualquality,
  title={VisualQuality-R1: Reasoning-Induced image quality sssessment via reinforcement learning to rank},
  author={Wu, Tianhe and Zou, Jian and Liang, Jie and Zhang, Lei and Ma, Kede},
  journal={arXiv preprint arXiv:2505.14460},
  year={2025}
}

@article{zhang2025vq,
  title={VQ-Insight: Teaching VLMs for AI-Generated video quality understanding via progressive visual reinforcement learning},
  author={Zhang, Xuanyu and Li, Weiqi and Zhao, Shijie and Li, Junlin and Zhang, Li and Zhang, Jian},
  journal={arXiv preprint arXiv:2506.18564},
  year={2025}
}

@article{shao2024deepseekmath,
  title={Deepseekmath: Pushing the limits of mathematical reasoning in open language models},
  author={Shao, Zhihong and Wang, Peiyi and Zhu, Qihao and Xu, Runxin and Song, Junxiao and Bi, Xiao and Zhang, Haowei and Zhang, Mingchuan and Li, YK and Wu, Yang and others},
  journal={arXiv preprint arXiv:2402.03300},
  year={2024}
}

@inproceedings{tsai2007frank,
  title={Frank: a ranking method with fidelity loss},
  author={Tsai, Ming-Feng and Liu, Tie-Yan and Qin, Tao and Chen, Hsin-Hsi and Ma, Wei-Ying},
  booktitle={Proceedings of the Annual International ACM SIGIR Conference on Research and Development in Information Retrieval},
  pages={383--390},
  year={2007}
}

@article{zhu2025internvl3,
  title={Internvl3: Exploring advanced training and test-time recipes for open-source multimodal models},
  author={Zhu, Jinguo and Wang, Weiyun and Chen, Zhe and Liu, Zhaoyang and Ye, Shenglong and Gu, Lixin and Tian, Hao and Duan, Yuchen and Su, Weijie and Shao, Jie and others},
  journal={arXiv preprint arXiv:2504.10479},
  year={2025}
}

@article{jia2024vqa,
  title={VQA$^2$: Visual question answering for video quality assessment},
  author={Jia, Ziheng and Zhang, Zicheng and Qian, Jiaying and Wu, Haoning and Sun, Wei and Li, Chunyi and Liu, Xiaohong and Lin, Weisi and Zhai, Guangtao and Min, Xiongkuo},
  journal={arXiv preprint arXiv:2411.03795},
  year={2024}
}

@article{jia2025scaling,
  title={Scaling-up perceptual video quality assessment},
  author={Jia, Ziheng and Zhang, Zicheng and Zhang, Zeyu and Liang, Yingji and Zhu, Xiaorong and Li, Chunyi and Han, Jinliang and Wu, Haoning and Wang, Bin and Zhang, Haoran and others},
  journal={arXiv preprint arXiv:2505.22543},
  year={2025}
}

@article{cao2025breaking,
  title={Breaking annotation barriers: Generalized video quality sssessment via ranking-based self-supervision},
  author={Cao, Linhan and Sun, Wei and Zhang, Kaiwei and Peng, Yicong and Zhai, Guangtao and Min, Xiongkuo},
  journal={arXiv preprint arXiv:2505.03631},
  year={2025}
}

@article{wu2023q,
  title={Q-bench: A benchmark for general-purpose foundation models on low-level vision},
  author={Wu, Haoning and Zhang, Zicheng and Zhang, Erli and Chen, Chaofeng and Liao, Liang and Wang, Annan and Li, Chunyi and Sun, Wenxiu and Yan, Qiong and Zhai, Guangtao and others},
  journal={arXiv preprint arXiv:2309.14181},
  year={2023}
}

@article{guo2025deepseek,
  title={Deepseek-r1: Incentivizing reasoning capability in llms via reinforcement learning},
  author={Guo, Daya and Yang, Dejian and Zhang, Haowei and Song, Junxiao and Zhang, Ruoyu and Xu, Runxin and Zhu, Qihao and Ma, Shirong and Wang, Peiyi and Bi, Xiao and others},
  journal={arXiv preprint arXiv:2501.12948},
  year={2025}
}

@article{li2018vmaf,
  title={VMAF: The journey continues},
  author={Li, Zhi and Bampis, Christos and Novak, Julie and Aaron, Anne and Swanson, Kyle and Moorthy, Anush and Cock, JD},
  journal={Netflix Technology Blog},
  volume={25},
  year={2018}
}

@article{mazurek2003role,
  title={A role for neural integrators in perceptual decision making},
  author={Mazurek, Mark E and Roitman, Jamie D and Ditterich, Jochen and Shadlen, Michael N},
  journal={Cerebral Cortex},
  volume={13},
  number={11},
  pages={1257--1269},
  year={2003},
  publisher={Oxford University Press}
}

@article{li2025perception,
  title={Perception, reason, think, and plan: A survey on large multimodal reasoning models},
  author={Li, Yunxin and Liu, Zhenyu and Li, Zitao and Zhang, Xuanyu and Xu, Zhenran and Chen, Xinyu and Shi, Haoyuan and Jiang, Shenyuan and Wang, Xintong and Wang, Jifang and others},
  journal={arXiv preprint arXiv:2505.04921},
  year={2025}
}

@article{li2022blindly,
  title={Blindly assess quality of in-the-wild videos via quality-aware pre-training and motion perception},
  author={Li, Bowen and Zhang, Weixia and Tian, Meng and Zhai, Guangtao and Wang, Xianpei},
  journal={IEEE Transactions on Circuits and Systems for Video Technology},
  volume={32},
  number={9},
  pages={5944--5958},
  year={2022},
  publisher={IEEE}
}

@inproceedings{wu2023towards,
  title={Towards explainable in-the-wild video quality assessment: a database and a language-prompted approach},
  author={Wu, Haoning and Zhang, Erli and Liao, Liang and Chen, Chaofeng and Hou, Jingwen and Wang, Annan and Sun, Wenxiu and Yan, Qiong and Lin, Weisi},
  booktitle={Proceedings of the ACM International Conference on Multimedia},
  pages={1045--1054},
  year={2023}
}

@inproceedings{duan2025finevq,
  title={Finevq: Fine-grained user generated content video quality assessment},
  author={Duan, Huiyu and Hu, Qiang and Wang, Jiarui and Yang, Liu and Xu, Zitong and Liu, Lu and Min, Xiongkuo and Cai, Chunlei and Ye, Tianxiao and Zhang, Xiaoyun and others},
  booktitle={Proceedings of the Computer Vision and Pattern Recognition Conference},
  pages={3206--3217},
  year={2025}
}

@inproceedings{zhang2025q,
  title={Q-Bench-Video: Benchmark the video quality understanding of LMMs},
  author={Zhang, Zicheng and Jia, Ziheng and Wu, Haoning and Li, Chunyi and Chen, Zijian and Zhou, Yingjie and Sun, Wei and Liu, Xiaohong and Min, Xiongkuo and Lin, Weisi and others},
  booktitle={Proceedings of the Computer Vision and Pattern Recognition Conference},
  pages={3229--3239},
  year={2025}
}

@inproceedings{wang2023exploring,
  title={Exploring clip for assessing the look and feel of images},
  author={Wang, Jianyi and Chan, Kelvin CK and Loy, Chen Change},
  booktitle={Proceedings of the AAAI Conference on Artificial Intelligence},
  volume={37},
  number={2},
  pages={2555--2563},
  year={2023}
}

@article{thomee2016yfcc100m,
  title={Yfcc100m: The new data in multimedia research},
  author={Thomee, Bart and Shamma, David A and Friedland, Gerald and Elizalde, Benjamin and Ni, Karl and Poland, Douglas and Borth, Damian and Li, Li-Jia},
  journal={Communications of the ACM},
  volume={59},
  number={2},
  pages={64--73},
  year={2016},
  publisher={ACM New York, NY, USA}
}

@article{mackin2018study,
  title={A study of high frame rate video formats},
  author={Mackin, Alex and Zhang, Fan and Bull, David R},
  journal={IEEE Transactions on Multimedia},
  volume={21},
  number={6},
  pages={1499--1512},
  year={2018},
  publisher={IEEE}
}

@inproceedings{chen2024internvl,
  title={Internvl: Scaling up vision foundation models and aligning for generic visual-linguistic tasks},
  author={Chen, Zhe and Wu, Jiannan and Wang, Wenhai and Su, Weijie and Chen, Guo and Xing, Sen and Zhong, Muyan and Zhang, Qinglong and Zhu, Xizhou and Lu, Lewei and others},
  booktitle={Proceedings of the IEEE/CVF Conference on Computer Vision and Pattern Recognition},
  pages={24185--24198},
  year={2024}
}

\clearpage
\maketitlesupplementary
\appendix

\begin{table*}[t]
    \centering
    {\fontsize{9}{10}\selectfont

    \renewcommand{\arraystretch}{0.8} 
    \resizebox{1\textwidth}{!}{
    \setlength{\tabcolsep}{4pt}
    \begin{tabular}{lccccccc}
        \toprule
        \textbf{Dataset} & \textbf{Year} & \textbf{\# of Videos} & \textbf{\# of Scenes} & \textbf{Resolution} & \textbf{Duration} & \textbf{Frame Rate} & \textbf{Distortion Type} \\
        \midrule
        \rowcolor{gray!20}
        \multicolumn{8}{l}{\textit{In-domain datasets for video quality scoring}} \\
        KoNViD-1k \cite{hosu2017konstanz} & 2017 & 1,200 & 1,200 & 540p & 8 & 24, 25, 30 & In-the-wild \\
        LIVE-VQC \cite{sinno2018large} & 2018 & 585 & 585 & 240p--1080p & 10 & 30 & In-the-wild \\
        YouTube-UGC \cite{wang2019youtube} & 2019 & 1,380 & 1,380 & 360p--4K & 20 & 30 & In-the-wild \\
        LSVQ \cite{ying2021patch} & 2021 & 38,837 & 38,837 & 99p--4K & 5--12 & $<$ 60 & In-the-wild \\
        \midrule
        \rowcolor{gray!20}
        \multicolumn{8}{l}{\textit{Out-of-distribution datasets for video quality scoring}} \\
        Waterloo-IVC-4K~\cite{li2019avc} & 2019 & 1,200 & 20 & 540p, 1080p, 4k & 10 & 24, 25, 30 & H.264 compression \\
        LIVE-YT-HFR  \cite{madhusudana2021subjective} & 2021 & 480 & 16 & 1080p & 6-10 & 24, 30, 60, 82, 98, 120 &  Frame rate, VP9 compression \\
        LIVE-YT-Gaming \cite{yu2022subjective} & 2022 & 600 & 600 & 360p--1080p & 8--9 & 30, 60 & PGC, UGC \\
        CGVDS  \cite{saha2023study} & 2023 & 360 & 15 & 480p, 720p, 1080p & 30 & 20, 30, 60 & H.264 compression \\
        VDPVE \cite{gao2023vdpve} & 2023 & 1,211 & 79 & Diverse & 8-15 & 24, 25, 30 & Enhancement \\
        \midrule
        \rowcolor{gray!20}
        \multicolumn{8}{l}{\textit{Datasets for video quality understanding}} \\
        FineVD~\cite{duan2025finevq} & 2025 & 6,104 & 6,104 & Diverse & 8 & Diverse & In-the-wild \\
        Q-Bench-Video~\cite{zhang2025q} & 2025 & 1,800 & 1,800 & Diverse & Diverse & Diverse & Diverse \\
        \bottomrule
    \end{tabular}}}
    \caption{An overview of our testing datasets.}
    \label{tab:dataset_summary}
\end{table*}

\section{Overview}

This supplementary material is organized into four sections. Section 1 provides detailed descriptions of all the testing benchmarks used in our experiments. Section 2 introduces the quality assessment methods used for comparison. Section 3 presents additional technical details of our model that were not covered in the main paper, including training hyper-parameters, model architecture, video preprocessing procedures, and the construction of temporal degradations. Finally, Section 4 offers further details of our experimental setup, including evaluation criteria, comprehensive ablation study results, and visualizations of our model's scoring and reasoning performance.

\section{More Details of Our Testing Benchmarks}

Table~\ref{tab:dataset_summary} provides an overview of our testing benchmarks, which encompass diverse content types, resolutions, durations, frame rates, and distortion types. In the following, we provide a detailed description of each dataset.

\begin{itemize}
    \item \textbf{KoNViD-1k~\cite{hosu2017konstanz}:} KoNViD-1k is an authentic video quality assessment (VQA) dataset consisting of 1,200 unique test videos with diverse real-world distortions. All videos are sampled from the YFCC100M dataset (Flickr)~\cite{thomee2016yfcc100m} based on a feature space including blur, colorfulness, contrast, spatial information, temporal information, and NIQE~\cite{mittal2012making}. Each video is clipped from the original source content and resized to 540p with a landscape layout. The frame rates of the videos are 24, 25, or 30 fps, and the duration of each video is 8 seconds.
    
    \item \textbf{LIVE-VQC~\cite{sinno2018large}:} LIVE-VQC is an authentic VQA dataset comprising 585 unique test videos with a wide range of real-world distortions. All videos are manually captured by multiple users using various mobile devices, resulting in diverse distortions such as camera motion artifacts and low-light conditions. The resolutions vary non-uniformly from 240p to 1080p, with both landscape and portrait orientations. The frame rates include 20, 24, 25, and 30 fps, and each video has a duration of 10 seconds.

    \item \textbf{YouTube-UGC~\cite{wang2019youtube}:} YouTube-UGC is an authentic VQA dataset consisting of 1{,}380 unique test videos (1{,}020 currently available) exhibiting a wide range of real-world distortions. The videos are sampled from YouTube using a feature space defined by spatial, color, temporal, and chunk-level variations, covering diverse content types such as HDR, screen recordings, animations, and gaming videos. The resolutions include 4K, 1080p, 720p, 480p, and 360p, with both landscape and portrait orientations. Frame rates range from 15 to 60 fps (including 15, 20, 24, 25, 30, 50, and 60), and each video lasts 20 seconds.

    \item \textbf{LSVQ~\cite{ying2021patch}:} LSVQ is a large-scale authentic VQA dataset comprising 38{,}837 unique videos with diverse real-world distortions. The videos span various resolutions and frame rates, with durations ranging from 5 to 12 seconds. The dataset is officially split into a training set of 28{,}056 videos and a testing set of 7{,}220 videos. Additionally, a dedicated 1080p testing subset containing 3{,}561 videos is provided for evaluating high-resolution scenarios.

    \item \textbf{Waterloo-IVC-4K~\cite{li2019avc}:} Waterloo-IVC-4K is a high-resolution VQA dataset that includes 20 pristine UHD video sequences (3840$\times$2160), each with a duration of 10 seconds. The source videos are carefully selected to cover a wide range of content types, including humans, plants, natural scenes, architectures, and computer-generated scenarios. Each source video is encoded using five representative video codecs (AVC, HEVC, VP9, AVS2, and AV1) at three spatial resolutions (3840$\times$2160, 1920$\times$1080, and 960$\times$540) and four different severity levels. In total, the dataset contains 1,200 distorted videos. 
    
    \item \textbf{LIVE-YT-HFR~\cite{madhusudana2021subjective}:} LIVE-YT-HFR is a VQA dataset designed to study the perceptual impact of frame rate and compression on video quality. It includes 16 source videos and 480 distorted videos, generated by applying six different frame rates and five levels of VP9 compression (including one lossless and four CRF levels). Among them, 11 sequences are sourced from the BVI-HFR dataset~\cite{mackin2018study}, with a resolution of 1920$\times$1080 and a duration of 10 seconds. The remaining five sequences are high-motion sports content captured by the Fox Media Group, presented in 3840$\times$2160 resolution, with video durations ranging from 6 to 8 seconds.

    \item \textbf{LIVE-YT-Gaming~\cite{yu2022subjective}:} LIVE-YT-Gaming is a VQA dataset specifically curated for gaming content. It consists of 600 short clips sourced from both Professionally-Generated Content (PGC) and User-Generated Content (UGC). The videos cover a variety of game genres and styles, with resolutions ranging from 360p to 1080p. All sequences are clipped to durations of 8 to 9 seconds.

    \item \textbf{CGVDS~\cite{saha2023study}:} CGVDS is a mobile cloud gaming VQA dataset consisting of 15 pristine gaming videos and 360 distorted videos generated by varying levels of H.264 compression. The videos cover multiple spatial resolutions, including 480p, 720p, and 1080p, with a fixed duration of 30 seconds.

    \item \textbf{VDPVE~\cite{gao2023vdpve}:} VDPVE is a VQA dataset constructed for Perceptual Video Enhancement. It contains a total of 1{,}211 enhanced video videos, divided into three sub-datasets: (1) 600 videos with color, brightness, and contrast enhancements; (2) 310 videos with deblurring enhancements; and (3) 301 videos with deshaking enhancements. In this study, we use an open-sourced subset of 839 videos from the training split to evaluate our model.

    \item \textbf{FineVD~\cite{duan2025finevq}:} FineVD is a large-scale fine-grained VQA dataset specifically designed for UGC content. It consists of 6{,}104 user-generated videos spanning various application scenarios, including on-demand and live-streaming services, general content, and short-form videos. A team of experts annotated each video from six quality dimensions—\textit{color}, \textit{noise}, \textit{artifact}, \textit{blur}, \textit{temporal}, and \textit{overall}. The dataset provides mean opinion scores (MOS) along with degradation type labels and quality descriptions, making it one of the most comprehensive fine-grained UGC VQA datasets to date.

    \item \textbf{Q-Bench-Video~\cite{zhang2025q}:} Q-Bench-Video is a benchmark specifically designed to systematically evaluate the video quality understanding capabilities of large multimodal models (LMMs). It covers a diverse range of video content, including natural scenes, AI-generated content (AIGC), and computer graphics. To ensure a balanced quality distribution, the benchmark samples videos with subjective annotations from multiple sources. It includes both multiple-choice questions and open-ended questions to evaluate model performance across varied quality assessment scenarios. In total, the benchmark comprises 1{,}800 videos and 2{,}378 annotated question-answer pairs for validation, offering a comprehensive framework for evaluating video quality understanding in LMMs.
\end{itemize}

\begin{table*}[!h]
\centering
{\fontsize{9}{10}\selectfont
    \resizebox{0.9\textwidth}{!}{
    \setlength{\tabcolsep}{16pt}

    \begin{tabular}{lcc}
    \toprule
    \textbf{Model Hyper-Parameters} & \textbf{Value} & \textbf{Description}\\
    \midrule
    \rowcolor{gray!20}
    \multicolumn{3}{l}{\textit{Parameters control data preprocessing}} \\
    max\_prompt\_length & 512 & / \\
    num\_generations & 4 & / \\
    max\_completion\_length & 2048 & / \\
    deepspeed stage & 2 & / \\
    Numerical precision & bfloat16 & / \\
    \rowcolor{gray!20}
    \multicolumn{3}{l}{\textit{Parameters control generation}} \\
    temperature & 0.9 & \textit{Temperature for sampling.}\\
    top\_p & 1.0 & \textit{Float that controls the cumulative probability of the top tokens to consider.}\\
    top\_k & 50 & \textit{Number of highest probability vocabulary tokens to keep for top-k-filtering.}\\
    \rowcolor{gray!20}
    \multicolumn{3}{l}{\textit{Parameters control training}} \\
    learning\_rate & 1e-6 & /\\
    $\beta$ & 0.04 & \textit{KL coefficient}\\
    epsilon & 0.2 & \textit{Epsilon value for clipping.} \\
    gradient\_accumulation\_steps & 2 & / \\
    num\_train\_epochs & 3 & /  \\
    batch\_size & 64 & / \\
    
    \bottomrule
    \end{tabular}}}
\caption{Details of the model hyper-parameters for training.}
\label{tab:model_hyperparams}
\end{table*}

\section{More Details of Our Comparing Methods}

\subsection{Unsupervised Methods}

Unsupervised image and video quality assessment methods aim to evaluate perceptual quality without relying on any labeled training data. All the methods described in this subsection are completely blind no-reference (NR) approaches that operate without any supervised learning.

For image-based methods, we uniformly sample one frame per second from each video, compute per-frame quality scores, and average them to obtain the final video-level score.

\begin{itemize}
    \item \textbf{NIQE~\cite{mittal2012making}:} NIQE constructs a ``quality aware'' collection of statistical features derived from a natural scene statistics (NSS) model. These features are extracted from a corpus of undistorted natural images and fit to a multivariate Gaussian (MVG) model. The quality of a test image is then quantified as the distance between the MVG of its NSS features and the MVG fitted on natural images.

    \item \textbf{VIIDEO~\cite{mittal2015completely}:} VIIDEO constructs perceptually relevant space-time statistical models based on temporal frame differences and analyzes deviations from these models using NSS. The model fits the NSS features to a MVG, and the video quality is inferred from the extent of deviation.

    \item \textbf{STEM~\cite{kancharla2021completely}:} STEM hypothesizes that natural videos follow a straighter temporal trajectory in the perceptual space, and that distortions introduce curvature. To quantify temporal quality, STEM computes the prediction error when linearly interpolating current frames from past ones in the HVS-transformed space. For spatial quality, STEM uses weighted NIQE scores of both current and previous frames. A simple linear combination of the temporal and spatial scores produces the final quality prediction.

    \item \textbf{CLIP-IQA~\cite{wang2023exploring}:} CLIP-IQA leverages the intrinsic perceptual capabilities and semantic priors encoded in CLIP without requiring any training on quality labels. It formulates IQA as a vision-language matching problem through prompt engineering, where positive and negative quality prompts (e.g., “Good photo.” vs. “Bad photo.”) are constructed to evaluate visual quality. To mitigate ambiguity in natural language interpretation, a contrasting prompt pair strategy is employed to ensure robust and consistent comparisons.
    
\end{itemize}

\subsection{Supervised Methods}
\begin{itemize}
    \item \textbf{SimpleVQA~\cite{sun2022deep}:} SimpleVQA introduces a lightweight yet effective framework for VQA. It employs an end-to-end spatial feature extractor trained on raw video pixels to capture quality-aware representations, while motion features are extracted from densely sampled low-resolution frames to efficiently model temporal distortions. A simple multilayer perceptron (MLP) is used to predict chunk-level quality scores, followed by temporal average pooling to obtain the final video-level score. To ensure robustness across resolutions, a multi-scale quality fusion strategy is adopted based on the contrast sensitivity function of the human visual system.

    \item \textbf{FAST-VQA~\cite{wu2022fast}:} FAST-VQA proposes a novel Grid Mini-patch Sampling (GMS) strategy that preserves local quality by sampling patches at the original resolution, while also capturing global quality via uniformly sampled mini-patches. These mini-patches are spliced and temporally aligned to form fragments. To process these fragments, the authors design a Fragment Attention Network (FANet), which effectively extracts quality-aware features from videos. By integrating GMS and FANet, FAST-VQA enables efficient end-to-end VQA with strong representational capability.

    \item \textbf{DOVER~\cite{wu2023exploring}:} DOVER extends FAST-VQA by incorporating a dual-branch architecture: a technical branch to capture low-level distortions and an aesthetic branch to assess high-level semantic composition, which reflects user preferences and content recommendation relevance. By disentangling these two complementary aspects, DOVER provides a more human-aligned and interpretable framework for VQA.

    \item \textbf{MinimalisticVQA~\cite{sun2024analysis}:} MinimalisticVQA (IX) utilizes the Swin Transformer-B~\cite{liu2022video}, pre-trained on ImageNet-1K~\cite{deng2009imagenet}, as a spatial quality analyzer to extract quality-aware features from key frames. To account for motion-related distortions, it integrates a temporal quality analyzer based on the SlowFast network~\cite{feichtenhofer2019slowfast}, which is pre-trained on the Kinetics-400~\cite{carreira2017quo} dataset. This dual-module design enhances the model’s capability in capturing both spatial details and temporal quality variations.
    
    \item \textbf{Q-Align~\cite{wu2024q}:} Q-Align introduces a novel training paradigm for LMM-based VQA by replacing direct numerical regression with discrete, text-based rating categories (e.g., ``excellent", ``good", ``fair", ``poor", ``bad") as supervision targets. During inference, the model computes the log-probabilities of each rating level, applies softmax normalization to obtain a probability distribution, and derives the final predicted quality score as the weighted average of the discrete labels.

    \item \textbf{VQA$^2$~\cite{jia2024vqa}:} VQA$^2$ is a series of LMMs that aim to unify low-level video quality assessment with high-level visual understanding. Built upon LLaVA-OneVision-Chat-7B~\cite{li2024llava}, VQA$^2$ models integrate both visual and motion tokens to capture fine-grained spatio-temporal quality details. The series includes two representative models: \textbf{VQA$^2$-Scorer} and \textbf{VQA$^2$-Assistant}. The VQA$^2$-Scorer is fine-tuned using full supervision to predict quality scores for user-generated and streaming videos. On top of this, the VQA$^2$-Assistant undergoes an additional training stage with quality-related instruction question-answer pairs, enabling it to perform more nuanced quality reasoning and understanding tasks while retaining strong scoring ability.
    
\end{itemize}

\subsection{RL-based Methods}
\begin{itemize}

    \item \textbf{Q-Insight~\cite{li2025q}:} Q-Insight adopts a multi-task reinforcement learning (RL) framework based on Group Relative Policy Optimization (GRPO), jointly optimizing two tasks: score regression and degradation perception, using only limited human ratings and distortion labels. The model generates groups of answers with explicit reasoning steps, each evaluated via dedicated reward functions: the score regression reward measures the alignment between predicted scores and MOS using a binary comparison mechanism, while the degradation perception reward assesses the accuracy of predicted distortion types and levels.

    \item \textbf{VisualQuality-R1~\cite{wu2025visualquality}:} VisualQuality-R1 employs GRPO to assess image quality by forming image pairs and calculating the difference between their predicted scores to derive a continuous fidelity~\cite{tsai2007frank} reward under the Thurstone model~\cite{thurstone2017law}. This reward signal is used to supervise the pairwise ranking optimization. By leveraging continuous score differences rather than discrete labels, the model achieves more fine-grained quality discrimination.

    \item \textbf{VQ-Insight~\cite{zhang2025vq}:} VQ-Insight employs GRPO for AIGC VQA. It adopts a progressive learning strategy: the model is first warmed up with image-level scoring tasks to establish basic quality perception, and then further optimized on video inputs using a set of multi-dimensional rewards, including temporal modeling, length control, score regression, and preference comparison. These rewards collaboratively enhance the model’s spatio-temporal quality understanding and alignment with human evaluation.

\end{itemize}

\section{More Details of VQAThinker Model}

\subsection{Model Hyper-parameters}
The detailed model training hyper-parameters are provided in Table~\ref{tab:model_hyperparams}.

\subsection{Details of Model Architecture}

\begin{figure}[!h]
\centering
\centerline{\epsfig{figure=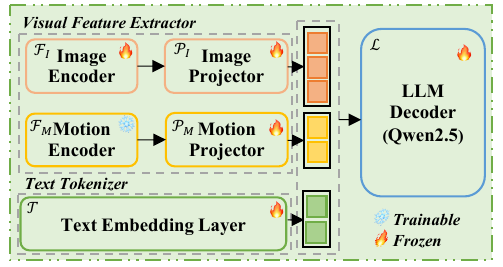,width=9cm}}
\caption{The structure of our LMM.}
\label{fig:lmm}
\end{figure}

As illustrated in Figure~\ref{fig:lmm}, our model mainly comprises three components: a visual feature extractor, a text tokenizer, and an LLM decoder.

\noindent\textbf{Visual Feature Extractor.}
Our visual feature extractor adopts a dual-branch design that processes sampled key frames through both spatial and temporal branch. Given a video $\bm{v}$, The spatial branch utilizes an image encoder (InternViT-300M-448px-V2\_5~\cite{chen2024internvl}) $\mathcal{F}_I$ to extract static visual features from individual sampled frames $\bm{v}_s$, while the temporal branch employs a pre-trained motion encoder (SlowFast~\cite{feichtenhofer2019slowfast}) $\mathcal{F}_M$ to capture motion-related dynamics across sampled frames. Both branches apply dedicated projection layers $\mathcal{P}_I$ and $\mathcal{P}_M$ to map their respective outputs into the language-aligned space:

\begin{equation}
f^s = \mathcal{P}_I(\mathcal{F}_I(\bm{v}_s)), \quad
f^t = \mathcal{P}_M(\mathcal{F}_M(\bm{v}_s)),
\end{equation}
The separated feature streams $f^s$ and $f^t$ allow the model to reason over both spatial and temporal aspects in a disentangled manner.

\noindent\textbf{Feature Fusion via the LLM.}
The input prompt $\bm{p}$ is first encoded into token embeddings $\bm{f}^p = \mathcal{T}(\bm{p})$ using tokenizer $\mathcal{T}$. Then, the visual features of the video (including frame features $\bm{f}^s$ and motion features $\bm{f}^t$) are injected into corresponding token positions. The fused sequence is passed to a pretrained LLM decoder for multimodal reasoning and quality prediction.
The model generates an output response $\bm{r}$, where the reasoning process is enclosed in \texttt{<think>}~\texttt{</think>} tags, and the final quality score is enclosed in \texttt{<answer>}~\texttt{</answer>} tags:
\begin{equation}
\begin{aligned}
\bm{r} &= \mathcal{L}(\bm{f}^s,\bm{f}^t,\bm{f}^p).
\end{aligned}
\end{equation}

\subsection{Details of Video Preprocessing}

During training, we uniformly sample 6 frames from each video and resize them to a resolution of $448 \times 448$ pixels. This design balances visual representation learning and computational efficiency. During inference, we increase the number of sampled frames to 12 and resize them to $560 \times 560$ pixels. This higher spatial and temporal resolution allows the model to access more comprehensive visual information, thereby improving its ability to understand and reason about video content under evaluation settings.

Specifically, for motion feature extraction, given the sampled frames $\bm{v}_s = \{v_i\}_{i=0}^{N-1}$ containing $N$ frames, where each $v_i$ denotes the $i$-th frame in the sequence, we first partition them into $N_c = \lfloor N / 3 \rfloor$ continuous chunks $\{\bm{c}_k\}_{k=0}^{N_c-1}$. For each chunk, the slow pathway features $\bm{f}^{\text{slow}}_k$ are extracted from the first frame, while the fast pathway features $\bm{f}^{\text{fast}}_k$ are computed over all sampled frames. Finally, the two types of features are concatenated along the channel dimension to obtain the final temporal feature representation $\bm{f}^t$. It is worth noting that these motion features only capture \textit{local temporal patterns} within the sampled frames.

\begin{figure*}[!h]
\centering
\centerline{\epsfig{figure=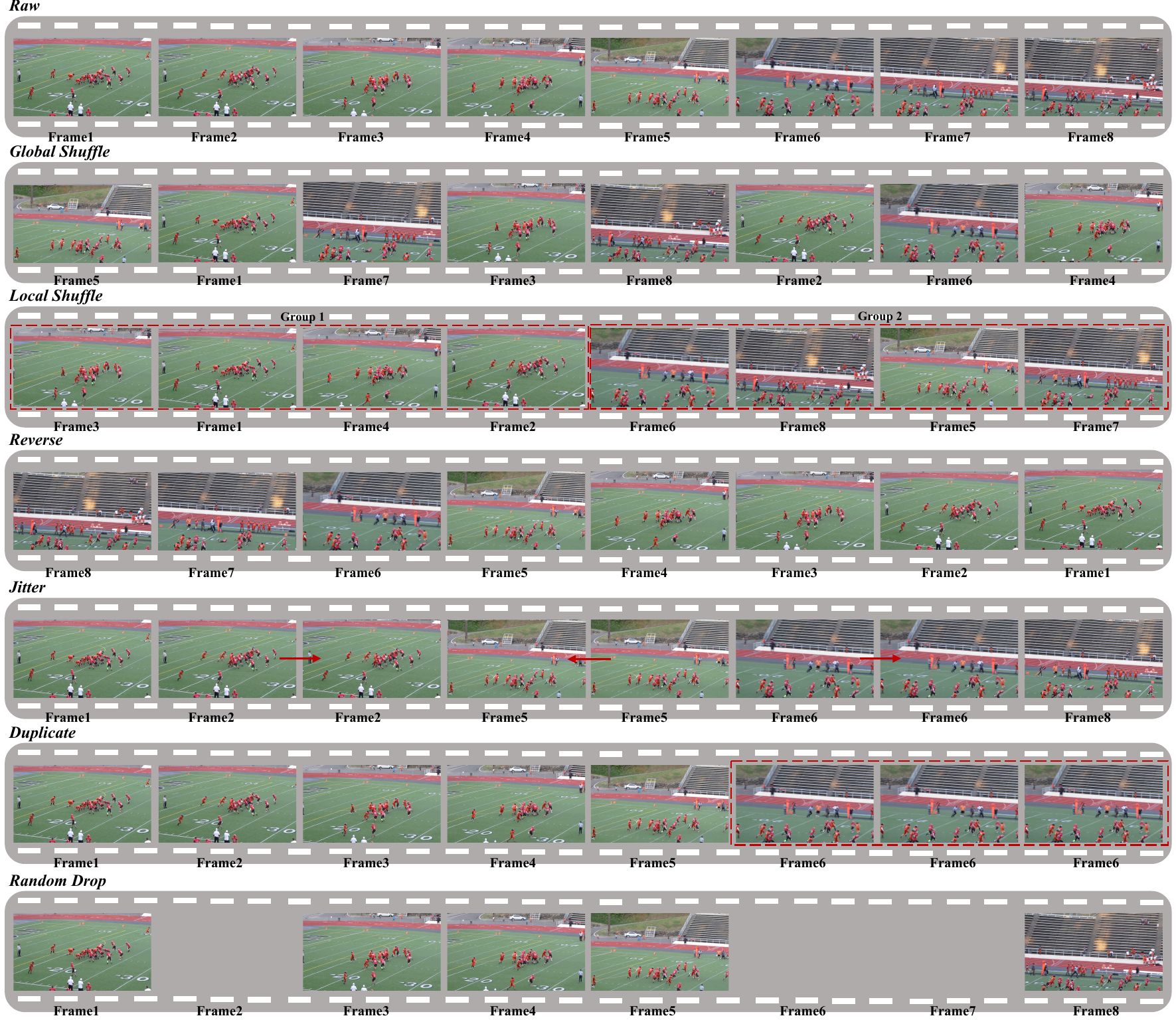,width=18cm}}
\caption{Illustration of different temporal perturbation modes.}
\label{fig:temp}
\end{figure*}

\subsection{Details of Temporal Degradation}

To simulate realistic temporal distortions that may occur in real-world video transmission or rendering, we apply a series of frame-level temporal perturbations to the input video $\bm{v}$, resulting in a degraded counterpart $\bm{v}_{\text{temp}}$. These perturbations aim to mimic common artifacts such as frame repetition, jitter, drop, and order disturbance. Specifically, , as illustrated in Figure~\ref{fig:temp}, we consider the following perturbation modes:

\begin{table*}[t]
    \centering
    {\fontsize{9}{10}\selectfont 
    \resizebox{1.0\textwidth}{!}{
    \renewcommand{\arraystretch}{1.0} 
    \setlength{\tabcolsep}{10pt}
    \begin{tabular}{c cc cc cc cc cc cc}
        \toprule
        \textbf{In-domain Datasets} & \multicolumn{2}{c}{\textbf{LSVQ$_\text{test}$}} & \multicolumn{2}{c}{\textbf{LSVQ$_\text{1080p}$}} &\multicolumn{2}{c}{\textbf{KoNViD-1k}} & \multicolumn{2}{c}{\textbf{LIVE-VQC}} & \multicolumn{2}{c}{\textbf{YouTube-UGC}} & \multicolumn{2}{c}{\textbf{Overall}}\\
        \cmidrule(lr){1-1} \cmidrule(lr){2-3} \cmidrule(lr){4-5} \cmidrule(lr){6-7} \cmidrule(lr){8-9} \cmidrule(lr){10-11} \cmidrule(lr){12-13} 
        \textbf{\# of videos} & \multicolumn{2}{c}{\textbf{7,182}} & \multicolumn{2}{c}{\textbf{3,573}} & \multicolumn{2}{c}{\textbf{1,200}} & \multicolumn{2}{c}{\textbf{585}} & \multicolumn{2}{c}{\textbf{1,020}} & \multicolumn{2}{c}{\textbf{---}} \\
        \cmidrule(lr){1-1} \cmidrule(lr){2-3} \cmidrule(lr){4-5} \cmidrule(lr){6-7} \cmidrule(lr){8-9} \cmidrule(lr){10-11} \cmidrule(lr){12-13} 
        \textbf{Methods}  & \textbf{SRCC} & \textbf{PLCC} & \textbf{SRCC} & \textbf{PLCC} & \textbf{SRCC} & \textbf{PLCC} & \textbf{SRCC} & \textbf{PLCC} & \textbf{SRCC} & \textbf{PLCC}  & \textbf{SRCC} & \textbf{PLCC} \\
        \midrule
         \textbf{SFT}  & 0.865 & 0.863 & 0.782 & 0.823 & 0.860 & 0.862 & 0.781 & 0.818 & 0.834 & 0.845 & 0.837 & 0.849 \\
         \textbf{GRPO} & 0.883 & 0.880 & 0.798 & 0.834 & 0.881 & 0.884 & 0.808 & 0.847 & 0.860 & 0.863 & 0.855 & 0.866 \\
        \midrule
        \textbf{Out-of-distribution Datasets} & \multicolumn{2}{c}{\textbf{LIVE-YT-Gaming}} & \multicolumn{2}{c}{\textbf{CGVDS}} & \multicolumn{2}{c}{\textbf{LIVE-YT-HFR}} & \multicolumn{2}{c}{\textbf{Waterloo-IVC-4K}} & \multicolumn{2}{c}{\textbf{VDPVE}} & \multicolumn{2}{c}{\textbf{Overall}} \\
        \cmidrule(lr){1-1} \cmidrule(lr){2-3} \cmidrule(lr){4-5} \cmidrule(lr){6-7} \cmidrule(lr){8-9} \cmidrule(lr){10-11} \cmidrule(lr){12-13} 
        \textbf{\# of videos} & \multicolumn{2}{c}{\textbf{600}} & \multicolumn{2}{c}{\textbf{357}} & \multicolumn{2}{c}{\textbf{480}} & \multicolumn{2}{c}{\textbf{1,200}} & \multicolumn{2}{c}{\textbf{839}} & \multicolumn{2}{c}{\textbf{---}} \\
        \cmidrule(lr){1-1} \cmidrule(lr){2-3} \cmidrule(lr){4-5} \cmidrule(lr){6-7} \cmidrule(lr){8-9} \cmidrule(lr){10-11} \cmidrule(lr){12-13} 
        \textbf{Methods} & \textbf{SRCC} & \textbf{PLCC} & \textbf{SRCC} & \textbf{PLCC} & \textbf{SRCC} & \textbf{PLCC} & \textbf{SRCC} & \textbf{PLCC} & \textbf{SRCC} & \textbf{PLCC}  & \textbf{SRCC} & \textbf{PLCC} \\
        \midrule
         \textbf{SFT} & 0.696  & 0.743 & 0.817 & 0.819 & 0.344 & 0.403 & 0.460 & 0.504 & 0.628 & 0.639 & 0.562 & 0.596 \\
         \textbf{GRPO} & 0.767 & 0.806 & 0.856 & 0.845 & 0.528 & 0.610 & 0.573 & 0.624 & 0.706 & 0.716 & 0.661 & 0.698 \\
        \bottomrule
    \end{tabular}
    }}
    \caption{Performance comparison between our model structures trained with SFT and GRPO strategies.}
    \label{tab:sftvsgrpo}
    \vspace{-0.4cm}
\end{table*}

\begin{itemize}

    \item \textbf{Global Shuffle:} To simulate extreme temporal disorder, we randomly permute all video frames, completely disrupting the original temporal sequence. Formally, given an input video $\bm{v} = \{f_1, f_2, \dots, f_T\}$ consisting of $T$ frames, the globally shuffled version $\bm{v}_{\text{shuffle}}$ is constructed by applying a random permutation $\pi$ over the frame indices:
    
    \[
    \bm{v}_{\text{shuffle}} = \{f_{\pi(1)}, f_{\pi(2)}, \dots, f_{\pi(T)}\}, \quad \pi \in \mathcal{S}_T,
    \]
    where $\mathcal{S}_T$ denotes the symmetric group of all possible permutations over $T$ elements.
    
    This operation completely destroys the global temporal continuity and disrupts any motion-related consistency across frames. It is designed to evaluate the model's ability to perceive video quality under severe temporal disorder.

    \item \textbf{Local Shuffle:} To simulate localized temporal corruption, we divide the input video into $N = \lfloor T / w \rfloor$ non-overlapping windows of fixed size $w$ (e.g., $w=4$). Each window is defined as:
    
    \[
    W_i = \{f_{(i-1)w + 1}, \dots, f_{iw}\}, \quad i = 1, 2, \dots, N.
    \]
    For each window $W_i$, we apply a local permutation $\pi_i \in \mathcal{S}_w$ to randomly shuffle the order of frames within that segment:
    \[
    W_i^{\text{shuffle}} = \{f_{\pi_i(1)}, f_{\pi_i(2)}, \dots, f_{\pi_i(w)}\}.
    \]
    The resulting shuffled video becomes:
    \[
    \bm{v}_{\text{local}} = \bigcup_{i=1}^{N} W_i^{\text{shuffle}}.
    \]
    
    This transformation introduces localized temporal inconsistency while preserving coarse-grained global structure. It aims to evaluate the model's ability to detect subtle quality degradations caused by short-range motion inconsistencies.

    \item \textbf{Reverse:} To simulate temporal distortion that disrupts natural motion patterns, we reverse the entire frame sequence of the video. Given an input video $\bm{v}$ with $T$ frames, we construct a reversed counterpart $\bm{v}_{\text{rev}}$ as:
    
    \[
    \bm{v}_{\text{rev}} = \{f_T, f_{T-1}, \dots, f_1\}.
    \]
    
    This transformation preserves the spatial content of each frame but completely inverts the temporal progression of actions. It creates unnatural motion trajectories (e.g., walking backward, objects returning to original positions), which do not occur in real-world videos. Such reversed sequences are useful to assess a model's sensitivity to motion dynamics and temporal causality.

    \item \textbf{Jitter:} To simulate frame-level temporal instability caused by jittery camera motion or temporal aliasing, we introduce localized frame substitutions in the video. Specifically, for each time step $t$, the original frame $f_t$ is replaced with a randomly selected frame from its immediate neighborhood:
    
    \[
    f_t' = f_{t + \delta}, \quad \delta \in \{-1, 0, +1\},
    \]
    where $\delta$ is sampled independently for each frame and boundary conditions are handled by clamping.
    
    The resulting video sequence $\bm{v}_{\text{jitter}} = \{f_1', f_2', \dots, f_T'\}$ maintains general temporal structure but introduces micro-temporal disturbances, mimicking realistic noise such as accidental camera shakes or frame capture jitter. Such perturbation slightly distorts motion consistency without altering spatial fidelity. It helps evaluate whether models are robust to fine-grained temporal inconsistencies that may impact perceived smoothness.

    \item \textbf{Duplicate:} To simulate frame freezing or repetition artifacts caused by network buffering or encoder malfunction, we introduce a duplication-based temporal distortion. Specifically, given an input video sequence, we randomly select a frame $f_k$ and duplicate it $n$ times. These duplicates are inserted at a random position $p$ (where $1 \leq p \leq T$). To keep the sequence length unchanged, we randomly drop $n$ frames from positions outside the inserted segment:

    \[
    \bm{v}_{\text{dup}} = \{f_1, \dots, f_{p-1}, \underbrace{f_k, \dots, f_k}_{n~\text{copies}}, f_p, \dots, f_{T-n+1}\},
    \]
    
    This transformation induces unnatural temporal stuttering while maintaining global content consistency.

    \item \textbf{Random Drop:} A fixed number of frames are randomly selected and permanently dropped from the original sequence without replacement. This models frame loss caused by unstable network transmission or decoding failure. We randomly select $n$ distinct frame indices $\{i_1, \dots, i_n\}$ from $\{1, \dots, T\}$. The corrupted sequence is constructed by removing the selected frames:
    \[
    \bm{v}_{\text{drop}} = \{f_t ~|~ t \in \{1, \dots, T\},~ t \notin \{i_1, \dots, i_n\}\}.
    \]

\end{itemize}

For each original video, we randomly apply one of the temporal perturbation methods described above to construct its degraded counterpart, which is then used for computing the temporal consistency reward. This encourages the model to consistently rate $\bm{v}$ higher than $\bm{v}_{\text{temp}}$, thereby improving its robustness to temporal distortion.

\section{More Details of Experiments}

\subsection{Detailed Description of Evaluation Criteria}

\subsubsection{Visual Quality Understanding.}
For visual quality scoring, we adopt two widely used criteria to evaluate the performance of VQA models: Spearman Rank Correlation Coefficient (SRCC) and Pearson Linear Correlation Coefficient (PLCC). SRCC measures the prediction monotonicity, reflecting how well the predicted scores preserve the rank order of the ground-truth scores. PLCC measures the linear correlation between predicted scores and ground-truth scores.

Given a set of $n$ predictions $\{p_i\}_{i=1}^{n}$ and ground-truth scores $\{g_i\}_{i=1}^{n}$, the SRCC is computed as:

\begin{equation}
\text{SRCC} = 1 - \frac{6 \sum_{i=1}^{n} (r_i - s_i)^2}{n(n^2 - 1)},
\end{equation}
where $r_i$ and $s_i$ are the ranks of $p_i$ and $g_i$, respectively.

The PLCC is calculated as:

\begin{equation}
\text{PLCC} = \frac{\sum_{i=1}^{n} (p_i - \bar{p})(g_i - \bar{g})}{\sqrt{\sum_{i=1}^{n} (p_i - \bar{p})^2} \sqrt{\sum_{i=1}^{n} (g_i - \bar{g})^2}},
\end{equation}
where $\bar{p}$ and $\bar{g}$ are the means of the predicted and ground-truth scores, respectively.

Higher values of SRCC and PLCC indicate better model performance in terms of preserving human quality perception and linear accuracy. Both metrics range from $-1$ to $1$, with values closer to $1$ indicating stronger alignment with human subjective quality ratings.

\begin{table*}[t]
    \centering
    {\fontsize{9}{10}\selectfont
    \resizebox{1\textwidth}{!}{
    \setlength{\tabcolsep}{6pt}
    \begin{tabular}{ccc|c|c|cc|cc|cc|cc|cc|cc}
        \toprule
        \multicolumn{3}{c|}{\textbf{Reward}} & \multirow{2}{*}{\textbf{Motion}} & \multirow{2}{*}{\textbf{Reasoning}}  & 
        \multicolumn{2}{c|}{\textbf{LSVQ$_\text{test}$}} & \multicolumn{2}{c|}{\textbf{LSVQ$_\text{1080p}$}} &
        \multicolumn{2}{c|}{\textbf{KoNViD-1k}} & \multicolumn{2}{c|}{\textbf{LIVE-VQC}} &
        \multicolumn{2}{c|}{\textbf{YouTube-UGC}} & \multicolumn{2}{c}{\textbf{Overall}} \\
        \cmidrule(lr){1-3} 
        \cmidrule(lr){6-7} \cmidrule(lr){8-9} \cmidrule(lr){10-11}
        \cmidrule(lr){12-13} \cmidrule(lr){14-15} \cmidrule(lr){16-17}
        \textbf{Rank} & \textbf{Reg.} & \textbf{Temp.} &   &  &
        \textbf{SRCC} & \textbf{PLCC} & \textbf{SRCC} & \textbf{PLCC} &
        \textbf{SRCC} & \textbf{PLCC} & \textbf{SRCC} & \textbf{PLCC} &
        \textbf{SRCC} & \textbf{PLCC} & \textbf{SRCC} & \textbf{PLCC} \\
        \midrule

         \ding{52} &         &         &  \ding{52} & \ding{52} & 0.812 & 0.811 & 0.720 & 0.733 & 0.779 & 0.782 & 0.747 & 0.771 & 0.782 & 0.793 & 0.780 & 0.785 \\
                 & \ding{52} &         &  \ding{52} & \ding{52} &       0.858 &  0.853     &   0.749    &   0.789    &   0.848   &   0.858   &  0.781  &   0.821  &   0.835    &   0.832    &    0.823   & 0.834      \\
        \ding{52} & \ding{52} &         & \ding{52} & \ding{52} & 0.874 & 0.872 & 0.797 & 0.833 & 0.876 & 0.882 & 0.797 & 0.842 & 0.855 & 0.859 & 0.849 & 0.860 \\
          \ding{52} &  \ding{52} &  & & \ding{52} & 0.866 & 0.866 & 0.790 & 0.827 & 0.869 & 0.874 & 0.770 & 0.827 & 0.835 & 0.837 & 0.840 & 0.853\\
         \ding{52} & \ding{52} & \ding{52} &  \ding{52} &        & 0.870 & 0.862 & 0.791 & 0.819 & 0.866 & 0.865 & 0.787 & 0.831 & 0.847 & 0.845 & 0.844 & 0.848 \\
         \ding{52} & \ding{52} & \ding{52} & \ding{52} & \ding{52} & 0.883 & 0.880 & 0.798 & 0.834 & 0.881 & 0.884 & 0.808 & 0.847 & 0.860 & 0.863 & 0.855 & 0.866 \\
        \midrule
        \multicolumn{3}{c|}{\textbf{Reward}} & \multirow{2}{*}{\textbf{Motion}} & \multirow{2}{*}{\textbf{Reasoning}} & 
        \multicolumn{2}{c|}{\textbf{LIVE-YT-Gaming}} & \multicolumn{2}{c|}{\textbf{CGVDS}} & 
        \multicolumn{2}{c|}{\textbf{LIVE-YT-HFR}} & \multicolumn{2}{c|}{\textbf{Waterloo-IVC-4K}} & 
        \multicolumn{2}{c|}{\textbf{VDPVE}} & \multicolumn{2}{c}{\textbf{Overall}} \\
        \cmidrule(lr){1-3} 
        \cmidrule(lr){6-7} \cmidrule(lr){8-9} \cmidrule(lr){10-11}
        \cmidrule(lr){12-13} \cmidrule(lr){14-15} \cmidrule(lr){16-17}
        \textbf{Rank} & \textbf{Reg.} & \textbf{Temp.} &   &  &
        \textbf{SRCC} & \textbf{PLCC} & \textbf{SRCC} & \textbf{PLCC} &
        \textbf{SRCC} & \textbf{PLCC} & \textbf{SRCC} & \textbf{PLCC} &
        \textbf{SRCC} & \textbf{PLCC} & \textbf{SRCC} & \textbf{PLCC} \\
        \midrule

        \ding{52} &         &         & \ding{52} & \ding{52} & 0.573 & 0.624 & 0.626 & 0.671 & 0.421 & 0.463 & 0.381 & 0.426 & 0.635 & 0.636 & 0.506 & 0.541 \\
                 & \ding{52} &         & \ding{52} & \ding{52} &         0.723 &  0.767   &   0.813    &   0.820    &    0.526   &  0.524     &    0.508   &    0.564   &    0.680   &   0.698    &   0.620    & 0.652      \\
        \ding{52} & \ding{52} &         & \ding{52} & \ding{52} & 0.733 & 0.775 & 0.823 & 0.824 & 0.501 & 0.576 & 0.552 & 0.605 & 0.693 & 0.705 & 0.638 & 0.677 \\
        \ding{52} &  \ding{52} &  &   & \ding{52} &  0.681 &  0.734  & 0.764 & 0.773 & 0.457 & 0.510 & 0.527 & 0.579 & 0.718 & 0.721 & 0.614 & 0.650 \\
        \ding{52} & \ding{52} & \ding{52} &  \ding{52} &        & 0.722 & 0.758 & 0.807 & 0.829 & 0.487 & 0.561 & 0.535 & 0.567 & 0.694 & 0.707 & 0.627 & 0.660 \\
         \ding{52} & \ding{52} & \ding{52} & \ding{52} &\ding{52} & 0.767 & 0.806 & 0.856 & 0.845 & 0.528 & 0.610 & 0.573 & 0.624 & 0.707 & 0.716 & 0.662 & 0.698 \\
        \bottomrule
    \end{tabular}
    }}
    \caption{Detailed experimental results of ablation study.}
    \label{tab:ablation}
\end{table*}

\subsubsection{Visual Quality Understanding.} We describe below the evaluation protocols used on two datasets: \textbf{FineVD} and \textbf{Q-Bench-Video}.

For FineVD, which provides five types of degradation labels (\textit{color}, \textit{noise}, \textit{artifact}, \textit{blur}, \textit{temporal}) for each video, we evaluate the accuracy of distortion type identification across all LMMs. Specifically, for models such as Q-Insight, VisualQuality-R1, and our proposed VQAThinker, we require each model to generate an overall quality description for each video. We then assess whether the generated description explicitly mentions the types of distortions present in the video. For other LMMs, we follow the evaluation protocol proposed in the original dataset paper~\cite{duan2025finevq}, where the model is asked a series of binary (yes-or-no) questions regarding the presence of each degradation type. To ensure fair and consistent evaluation, all model outputs are assessed with the assistance of GPT-4o. We manually verified that the judgments made by the large language model are highly consistent with human annotations, and therefore adopt these results as the final accuracy scores.

For Q-Bench-Video, we evaluate the capability of LMMs to assess single videos, as each input sample in this task consists of a single video. According to the benchmark design, queries related to single videos are categorized into two types: (1) \textbf{Global perception}, which concerns the overall visual quality of the video, such as \textit{“How is the overall contrast of this video?”}; and (2) \textbf{Referring perception}, which focuses on the visual quality of specific elements within the video, like \textit{“What is the most apparent distortion when the player strikes the ball?”} Through these two types of queries, the benchmark comprehensively evaluates the models’ ability to perceive both the overall and localized aspects of video quality. All questions are presented in a multiple-choice format with 2 to 4 candidate answers, and we report the accuracy of the selected answers to quantify the model's performance.

\subsection{Detailed Ablation Study}

While the main paper reports the overall performance under in-domain and out-of-distribution (OOD) settings, we present here a more fine-grained ablation study on all 10 datasets. As shown in Table~\ref{tab:ablation}, each proposed component contributes meaningfully to the overall performance. Removing any individual module results in consistent performance degradation across most datasets, underscoring their complementary roles and the necessity of the full design.

Table \ref{tab:sftvsgrpo} presents a performance comparison of our model under two different training strategies: Supervised Fine-Tuning (SFT) and GRPO. In this comparison, all other settings are kept identical, with the only difference being the training strategy. As shown, GRPO consistently outperforms SFT across all datasets, demonstrating its strong ability to model video quality scoring tasks.

\begin{figure*}[!h]
\centering
\centerline{\epsfig{figure=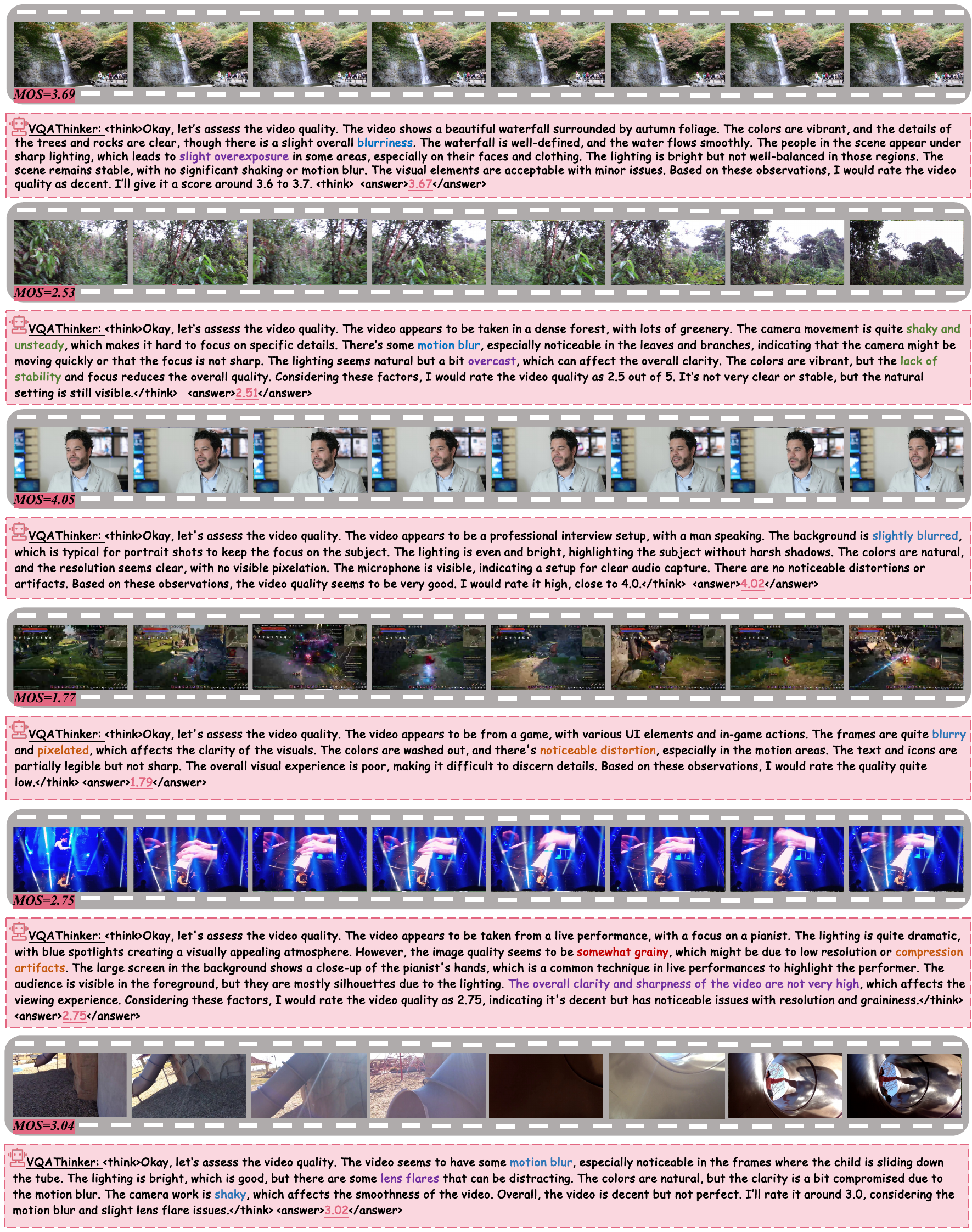,width=18cm}}
\caption{Examples illustrating our model's visual quality scoring and understanding capability. All MOS scores are linearly rescaled to a unified range of 1–5.}
\label{fig:pred}
\end{figure*}

\subsection{Performance Visualization}

To better illustrate the effectiveness of our method, we present qualitative examples from multiple test sets. These examples span a wide range of content categories and degradation types, including motion blur, compression artifacts, and color distortion, among others. As shown in Figure~\ref{fig:pred}, our model is capable of generating accurate and fine-grained quality scores that align well with human perception. Moreover, it demonstrates strong sensitivity to various distortions and degradations, reflecting a robust quality understanding ability across diverse visual scenarios. This further validates the model’s comprehensive capability in both quality estimation and perceptual alignment.

\end{document}